%% file: ms.tex
\documentclass[10pt,twocolumn,letterpaper]{article}
\pdfoutput=1

\usepackage{cvpr}              
\usepackage[accsupp]{axessibility}  
\usepackage{graphicx}
\usepackage{amsmath}
\usepackage{amssymb}
\usepackage{booktabs}
\usepackage{comment}
\usepackage{graphics}
\usepackage{amsmath}
\usepackage{diagbox}
\usepackage{multirow}

\usepackage[pagebackref,breaklinks,colorlinks]{hyperref}

\usepackage[capitalize]{cleveref}
\crefname{section}{Sec.}{Secs.}
\Crefname{section}{Section}{Sections}
\Crefname{table}{Table}{Tables}
\crefname{table}{Tab.}{Tabs.}

\def\cvprPaperID{1042} 
\def\confName{CVPR}
\def\confYear{2023}

\newcommand{\nerf}{NeRF\@\xspace}
\newcommand{\nerfs}{NeRFs\@\xspace}
\newcommand{\nerfrpn}{NeRF-RPN\@\xspace}

\DeclareMathOperator{\clip}{clip}

\begin{document}

\title{NeRF-RPN: A general framework for object detection in NeRFs}

\author{
\begin{tabular}{ccccc}
Benran Hu$^{1}$\thanks{ Equal contribution. The order of authorship was determined alphabetically. } & 
Junkai Huang$^{1}${\footnotemark[1]} & 
Yichen Liu$^{1}${\footnotemark[1]} & 
Yu-Wing Tai$^{1,2}$ &
Chi-Keung Tang$^{1}$ 
\end{tabular}
\\
\begin{tabular}{lr}
$^1$The Hong Kong University of Science and Technology &
$^2$Kuaishou Technology 
\end{tabular} \\
}

\maketitle

\renewcommand{\thefootnote}{\fnsymbol{footnote}}
\footnotetext[2]{This research is supported in part by the Research Grant Council of the Hong Kong SAR under grant no. 16201420.}

\newif\ifshowsolution

\input{0_abstract}
\input{1_intro}

\input{2_related}

\input{3_method}

\input{4_dataset}
\input{5_experiments}

\input{7_conclusions}

{\small
\bibliographystyle{ieee_fullname}
\bibliography{ref}
}

\end{document}


\title{NeRF-RPN: A general framework for object detection in NeRFs
\\Supplementary Material}

\author{
\begin{tabular}{ccccc}
Benran Hu$^{1}$\thanks{ Equal contribution. The order of authorship was determined alphabetically. } & 
Junkai Huang$^{1}${\footnotemark[1]} & 
Yichen Liu$^{1}${\footnotemark[1]} & 
Yu-Wing Tai$^{1,2}$ &
Chi-Keung Tang$^{1}$ 
\end{tabular}
\\
\begin{tabular}{lr}
$^1$The Hong Kong University of Science and Technology &
$^2$Kuaishou Technology 
\end{tabular} \\
}
\maketitle

\renewcommand{\thefootnote}{\fnsymbol{footnote}}
\footnotetext[2]{This research is supported in part by the Research Grant Council of the Hong Kong SAR under grant no. 16201420.}

\noindent {\em Please watch the supplementary video for visualizing our 3D region proposals on NeRFs at \url{https://youtu.be/M8_4Ih1CJjE}.}

\section{More Details on Dataset Construction}

\paragraph{Hypersim} As mentioned, we perform extensive cleaning based on the NeRF reconstruction quality. The number of camera poses on each trajectory in the Hypersim dataset is limited to 100, which is too sparse for NeRF training for many larger scenes, and usually produces fuzzy NeRF results strewn with a lot of dangling reconstruction errors or ``floaters'' to be removed. To remove these unsatisfactory scenes, we train NeRF models on all the scenes, and use a subset of training poses together with randomly interpolated poses as validation camera poses to examine the NeRF quality. We use the NeRF implementation from instant-NGP~\cite{mueller2022instant} and run at least 10k training iterations for each scene. By manually checking the NeRF rendering results, we filter out the following types of scenes: 1) scenes containing no objects; 2) scenes where  a significant number of object bounding boxes are missing; 3) scenes that are too blurry, or the objects which cannot be clearly separated from floaters. After cleaning the scenes, we further clean the object bounding boxes based on the criteria aforementioned in the paper.

\paragraph{3D-FRONT} We spent much effort to split complex scenes into individual rooms and cleaning up bounding boxes. In order to obtain data with suitable size for NeRF training, we first manually partition each scene into individual rooms according to the given layout of the scene. For each selected room, we generate 200$\sim$300 camera poses, including 100$\sim$150 general views randomly distributed in the room, and 15$\sim$20 close-up views for each object within the given room. With these poses, we use~\cite{denninger2019blenderproc} to render 2D images for NeRF training.

\paragraph{Read-World Data} SceneNN is a real-world indoor dataset with around 100 scenes, where RGB-D images with predicted poses, bounding boxes of objects and reconstructed meshes are provided for each scene. 
We first filter the images by choosing the image with highest sharpness (variance of Laplacian) among every 20 consecutive frames. Then, we project bounding boxes onto chosen images using camera poses to determine camera pose correctness and eliminate incorrect camera poses manually. 
A total of 16 scenes survive the above, and we 
use~\cite{mueller2022instant} to reconstruct them. 

\section{Ablation on NeRF Sampling Strategies}
To investigate how view-dependent radiance information from NeRF affects the performance of our method, we experiment the following sampling patterns:
\begin{enumerate}
\setlength{\itemsep}{-2pt}
    \item use density only;
    \item in addition to density, use the average radiance sampled from 18 fixed viewing directions in the form of $(\cos(\phi)\cos(\theta), \cos(\phi)\sin(\theta), \sin(\phi))$, where $\phi \in \{\frac{\pi}{3}, 0, -\frac{\pi}{3}\},\, \theta \in \{\frac{k\pi}{3} \,|\, k\in\mathbb{N}, 0\leq k \leq 5\}$;
    \item in addition to density, use the average radiance sampled from all training camera viewing directions;
    \item similar as 3) above, but only average from training camera views of which the viewing frustum contains the sample point. If a sample point is invisible in all frustums, we use the same scheme as~3) above;
    \item in addition to density, use the coefficients of the Spherical Harmonics (SH) at the sample point up to degree $l=3$. The SH function is fitted similarly as in \cite{yu2021plenoctrees} by uniformly sampling radiance from 300 directions on a sphere.
\end{enumerate}
\input{tabs/tab-ablation_sampling}
Table~\ref{tab:ablation_sampling} shows the results of different sampling methods above on the 3D-FRONT test set, using VGG19 as the backbone and the anchor-free RPN head. The results might be counter-intuitive as finely-curated radiance information impairs the performance. However, we speculate that density alone is sufficient for the region proposal task as it involves only a binary classification between objects and background which relies less on object semantics. Additionally, in this case, such extra radiance information may lead to more severe over-fitting and thus lower performance on a relatively small dataset such as 3D-FRONT. Nevertheless, the semantics carried in radiance information may be helpful for downstream classification tasks or the detection of secondary object structures.

\section{Objectness Classification}
As mentioned in Section 3.4, we implement a binary objectness classification model. We choose Swin-S\cite{liu2021Swin} as the backbone in the experiments and use the top 2,500 proposals from RPN after NMS with an IoU threshold of 0.3. We fine-tune the feature extractor trained on RPN with AdamW \cite{LoshchilovH19adamw}, an initial learning rate of 0.0001, and a weight decay of 0.0001. We set $\lambda = 5.0$ in the loss function and also adopt the same augmentation strategy in RPN training. During testing, we use ROIs with objectness scores larger than 0.5 to calculate the average precision (AP). Table ~\ref{tab:objectness} illustrates our results. We find that the APs do not increase as expected and we speculate that this results from the limited resolution of the feature volumes. The ROIs projected onto the coarser-level feature volumes can have similar or smaller sizes compared to our ROI pooling output, while a rotated interpolation over these low-resolution feature volumes can lead to high resampling errors and cannot produce precise features for each rotated ROI. Moreover, the quality of the NeRF models can also affect the ROI quality and the objectness classification performance. However, our objectness classification architecture might still be useful in many downstream tasks like object detection where the ROI features are required, especially when a higher-resolution feature pyramid is supplied. 
\input{tabs/tab-objectness}

\section{2D Projection}
\input{tabs/tab-2dproj_loss}
We project the 3D bounding boxes as aforementioned. 
However, we believe 3D features from NeRF already contain sufficient information for precise 3D bounding box regression, which renders the 2D projection loss redundant. This is corroborated by the results in Table~\ref{tab:proj_loss}, where introducing the extra loss does not help with performance. Therefore, we do not use 2D projection loss for other results presented in this paper. The 2D projection loss may however still be helpful when 3D supervision is unavailable.

\section{Video Results}
Please watch the supplementary video for  moving 3D demonstration of our test-time region proposal results and heat maps in various examples.

{\small
\bibliographystyle{ieee_fullname}
\bibliography{ref}
}

%% file: 0_abstract.tex
\begin{abstract}
This paper presents the first significant object detection framework, NeRF-RPN, which directly operates on NeRF. Given a pre-trained NeRF model, NeRF-RPN aims to detect all bounding boxes of objects in a scene. By exploiting a novel voxel representation that incorporates multi-scale 3D neural volumetric features, we demonstrate it is possible to regress the 3D bounding boxes of objects in NeRF directly without rendering the NeRF at any viewpoint. NeRF-RPN is a general framework and can be applied to detect objects without class labels. We experimented NeRF-RPN with various backbone architectures, RPN head designs and loss functions. All of them can be trained in an end-to-end manner to estimate high quality 3D bounding boxes. To facilitate future research in object detection for NeRF, we built a new benchmark dataset which consists of both synthetic and real-world data with careful labeling and clean up. Code and dataset are available at \url{https://github.com/lyclyc52/NeRF_RPN}.

\end{abstract}


%% file: 1_intro.tex
\section{Introduction}


3D object detection is fundamental to 
important applications such as robotics and autonomous driving, which require  scene understanding in 3D. Most existing relevant methods require 3D point clouds input or at least RGB-D images acquired from 3D sensors. Nevertheless, recent advances in Neural Radiance Fields (\nerf)~\cite{mildenhall2020nerf} provide an effective alternative approach to extract highly semantic features of the underlying 3D scenes from 2D multi-view images. Inspired by Region Proposal Network (RPN) for 2D object detection, in this paper, we present the first 3D 
\nerfrpn, which directly operates on the \nerf representation of a given 3D scene learned entirely from RGB images and camera poses. Specifically, given the radiance field and the density extracted from a NeRF model, our method  produces bounding box proposals, which can be deployed in downstream tasks. 

Recently, \nerf has provided very impressive results in novel view synthesis, 
while 3D object detection has become increasingly important in many real-world applications such as autonomous driving and augmented reality. Compared to 2D object detection, detection in 3D is more challenging due to the increased difficulty 
in data collection where various noises in 3D can be captured as well. Despite some good works, there is a lot of room for exploration in the field of 3D object detection. Image-based 3D object detectors either use a single image (e.g.,~\cite{brazil2019m3drpn, wang2019pseudolidar, chen20153dop}) or utilize multi-view consensus of multiple images (e.g.,~\cite{rukhovich2022imvoxelnet, wang2022detr3d, liu2022petr}). Although the latter use multi-view projective geometry to combine information in the 3D space, they still use 2D features to guide the pertinent 3D prediction. Some other 3D detectors based on point cloud representation (e.g.,~\cite{ zhou2018voxelnet, qi2018frustum, mao2021votr, liu2021groupfree}) heavily rely on accurate data captured by sensors. To our knowledge, there is still no representative work on direct 3D object detection in \nerf.

\input{figs/fig-teaser}
Thus, we propose \nerfrpn to propose 3D ROIs in a given \nerf representation. Specifically, the network takes as input the 3D volumetric information extracted from \nerf, and directly outputs 3D bounding boxes of ROIs. \nerfrpn will thus be a powerful tool for 3D object detection in \nerf by adopting the ``3D-to-3D learning'' paradigm, taking full advantages of 3D information inherent in \nerf and predicting 3D region proposals directly in 3D space.

As the first significant attempt to perform 3D object detection directly in NeRFs trained from multi-view images, this paper's focus contributions consist of:
\begin{itemize}
\setlength{\itemsep}{-2pt}
\item First significant attempt on introducing RPN to NeRF for 3D objection detection and related tasks. 

\item A large-scale public indoor \nerf dataset for 3D object detection, based on the existing synthetic indoor dataset Hypersim~\cite{roberts:2021} and 3D-FRONT~\cite{fu20213d}, and real indoor dataset ScanNet~\cite{dai2017scannet} and SceneNN~\cite{hua2016scenenn}, carefully curated for NeRF training. 
\item 
Implementation and comparisons of \nerf-RPNs on various backbone networks, detection heads and loss functions. Our model can be trained in 4 hrs using 2 NVIDIA RTX3090 GPUs. At runtime, it can process a given NeRF scene in 115 ms (excluding postprocessing) while achieving a 99\% recall on our 3D-FRONT \nerf dataset.
\item Demonstration of 3D object detection over NeRF and related applications based on our NeRF-RPN.

\end{itemize}

%% file: figs/fig-teaser.tex
\begin{figure}[t]
\centering
    \includegraphics[width=1\linewidth]{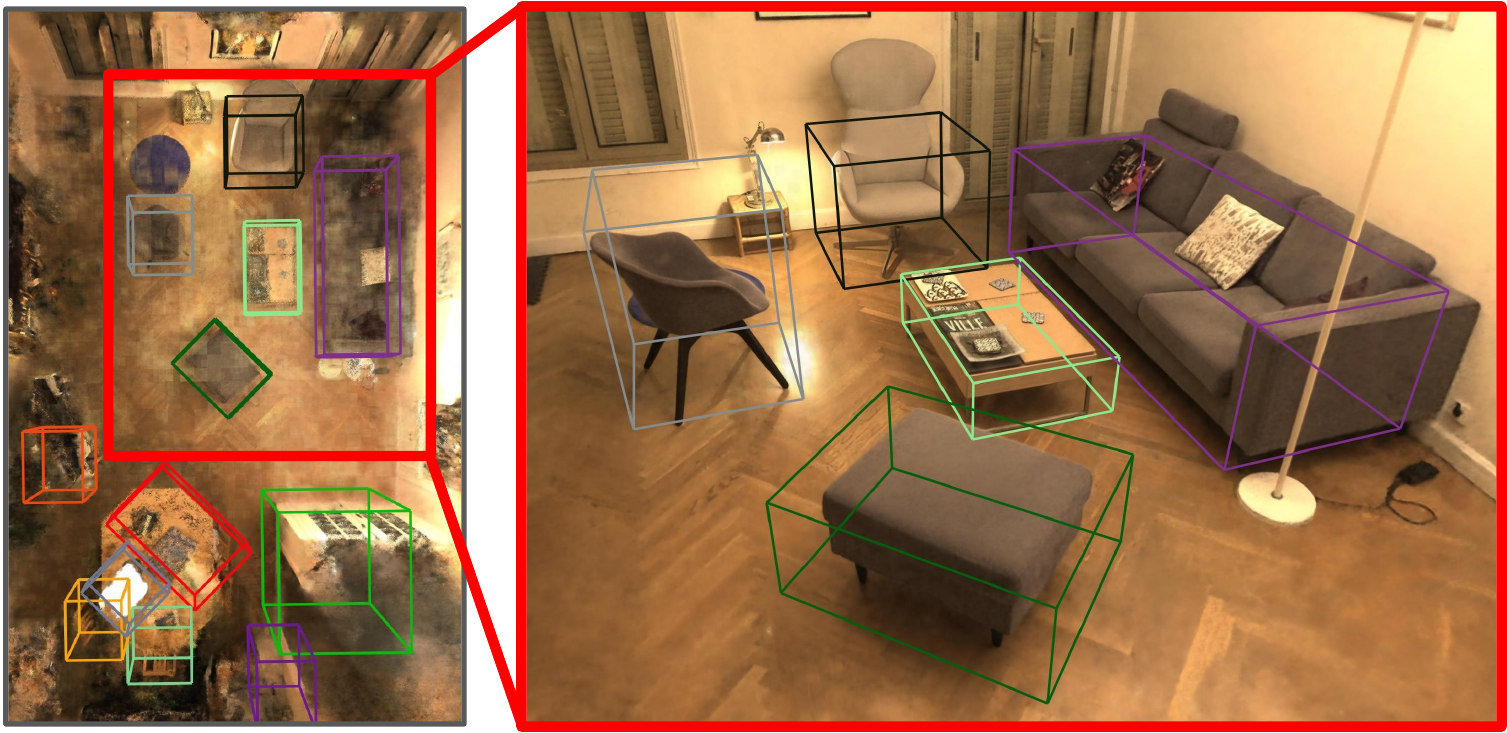}
    \vspace{-0.15in}
    \caption{\textbf{Region proposal results on a NeRF.} Top 12 proposals in eight orientations with highest confidence are visualized. The NeRF is trained from the \textit{Living Room} scene from INRIA \cite{inriadataset}.}\vspace{-0.15in}
    \label{fig:teaser}
\end{figure}

%% file: 2_related.tex
\section{Related Work}

\subsection{NeRF}
Neural radiance field (NeRF)\cite{mildenhall2020nerf} has become the mainstream approach for novel view reconstruction, which  models the geometry and appearance of a given scene in a continuous and implicit radiance field parameterized by an MLP. 
Following this work, instant neural graphics primitive~\cite{mueller2022instant} applies hash encoding to reduce the training time dramatically. PlenOctrees~\cite{yu2021plenoctrees} uses an octree-based radiance field and a grid of spherical basis functions to accelerate rendering and appearance decoding. TensoRF~\cite{Chen2022ECCV} projects a 3D point onto three 2D planes to encode the positional information. Although these works use different approaches to model structures, they achieve the same goal of taking as input \textup{xyz} coordinates and 3D camera poses to generate  view-dependent RGB color and  volume density at each position to render the images from a given view point. 
\nerf not only provides structural details of a 3D scene but is also conducive to 3D training, where only posed RGB images are required, thus making this representation also suitable for 3D object detection.


\subsection{Object Detection and Region Proposal Network}
Subsequent to~\cite{alexnet2012} and recent GPU advances, deep convolutional neural network (CNN) has become the mainstream 
approach for object detection given single images. Object detection 
based on deep learning can be divided into anchor-based methods and anchor-free methods. Anchor-based methods, including two-stage methods \cite{girshick2014rcnn, he2015sppnet, girshick2015fast, ren2015faster, he2017mask} and one-stage methods \cite{Redmon_2016_CVPR, Liu_2016, Lin_2017_ICCV, https://doi.org/10.48550/arxiv.1701.06659, https://doi.org/10.48550/arxiv.1712.00960, Zhang_2018_CVPR, Liu_2018_ECCV}, first generate a large number of preset anchors with different sizes and aspect ratios on a given image, then predict the labels and box regression offsets of these anchors. For two-stage methods, relatively coarse region proposals are first generated from anchors, followed by refining such coarse proposals and inferring the labels. 
In contrast to anchor-based methods, anchor-free methods \cite{Law_2018_ECCV, Duan_2019_ICCV, Dong_2020_CVPR, Zhou_2019_CVPR, https://doi.org/10.48550/arxiv.1904.07850, https://doi.org/10.48550/arxiv.1904.02948, Tian_2019_ICCV} predict on feature maps directly.

Region Proposal Network (RPN) was first introduced in~\cite{ren2015faster} to propose regions in an image that may contain objects for subsequent refinement. RPN uses shared convolutional layers to slide though local regions on the feature maps from feature extraction layers and feeds the transformed features into a box-regression head and a box-classification head in parallel. In~\cite{ren2015faster}, RPN is applied on the feature map from the last shared convolution layer only, whereas more recent works such as Feature Pyramid Networks (FPN)~\cite{Lin_2017_CVPR} utilize multi-scale feature maps. Our proposed method adapts the idea of sliding window from 2D RPN and also utilizes FPN in a 3D fashion.  

\subsection{3D Object Detection}

Based on the input form, current 3D object detectors can be categorized as point cloud-based and RGB-based methods. Many point cloud-based methods first transform point clouds into voxel forms to subsequently operate on the 3D feature volume through convolution \cite{song2016deep, zhou2018voxelnet, mao2021votr, rukhovich2021fcaf3d, gwak2020gsdn} or Transformers \cite{mao2021votr, vaswani2017attention}. However, the large memory footprint of the voxel representation constrains the resolution used. While sparse convolution~\cite{graham20183sparseconv} and 2D projection have been adopted to alleviate the issue, works directly operating on raw point clouds have been proposed recently~\cite{liu2021groupfree, wang2022rbgnet, misra20213detr, qi2018frustum, qi2017pointnet, qi2019votenet, lahoud20172ddriven, Shi_2019_CVPR_pointrcnn}. Most of them partition points into groups and apply classification and bounding box regression to each group. Criteria used for grouping include 3D frustums extruded from 2D detection~\cite{qi2018frustum}, 3D region proposals~\cite{Shi_2019_CVPR_pointrcnn, shi2021pvrcnn}, and voting~\cite{qi2019votenet, qi2020imvotenet, wang2022rbgnet}. GroupFree3D ~\cite{liu2021groupfree} and Pointformer~\cite{Pan_2021_CVPR}, on the other hand, use Transformers to attend over all points instead of grouping.

\input{figs/fig-main}

3D objection detection on single images or posed multi-view RGB images is more challenging and relatively less explored. Early attempts in monocular 3D objection detection first estimate the per-pixel depths~\cite{chen20153dop, xu2018mlfusion}, pseudo-LiDAR signals~\cite{wang2019pseudolidar, you2019pseudolidarplusplus, qian2020end2endpseudolidar}, or voxel information \cite{roddick2018orthographic} from an RGB image,  performing detection on the reconstructed 3D features.
Later works have extended 2D object detection methods to operate in 3D. For instance, M3D-RPN~\cite{brazil2019m3drpn} and MonoDIS~\cite{simonelli2019monodis} use 2D anchors for detection and predict a 2D to 3D transformation. FCOS3D~\cite{wang2021fcos3d} extends FCOS~\cite{tian2019fcos} to predict 3D information. More research recently has been focused on the multi-view case. ImVoxelNet~\cite{rukhovich2022imvoxelnet} projects 2D features back to a 3D grid and applies a voxel-based detector on it. DETR3D~\cite{wang2022detr3d} and PETR~\cite{liu2022petr} adopt similar designs as DETR~\cite{carion2020detr}, both trying to fuse 2D features and 3D position information. 
Although these image-based methods can assist the region proposal task in NeRF, they do not utilize the inherent 3D information from NeRF and are thus limited in their accuracy.

While it is possible to sample from NeRF and produce a voxel or point cloud representation on which previous 3D object detection methods can be applied, such conversion can be rather ad-hoc, depending on both the NeRF structure and the reconstruction quality. Noise and poor fine-level geometry approximation in these converted representations also pose challenges to existing 3D object detectors. Note that unlike point cloud samples, which cover only the surface (crust) of objects, the density in NeRF distributes over the interior as well. Clearly, existing methods fail to utilize this important solid object information, which is adequately taken into account by our \nerfrpn. Besides, there exists no 3D object detection dataset tailored for the NeRF representation, which also limits the advancement of 3D object detection in NeRF.



%% file: figs/fig-main.tex
\begin{figure*}[t]
    \centering
    \includegraphics[width=0.95\linewidth]{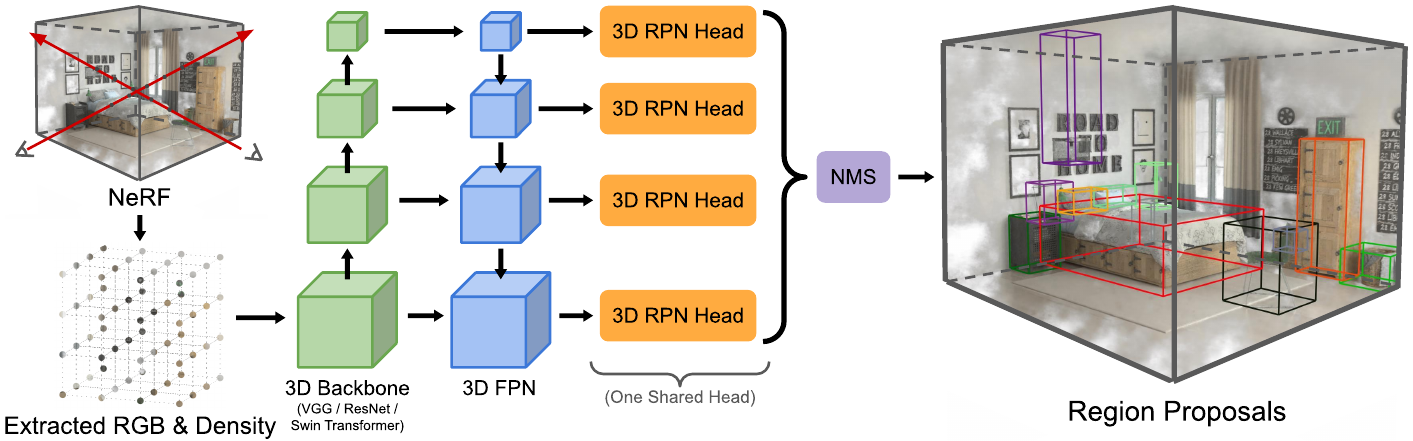}
    \vspace{-0.1in}
    \caption{\textbf{NeRF-RPN.} Our method first samples a grid of points in NeRF and extracts their RGB and density. The extracted volumetric features are then passed through a 3D backbone network to get deep multi-scale 3D features, which are fused with a 3D FPN and fed to the 3D RPN head to produce the region proposals.}\vspace{-0.2in}
    \label{fig:main}
    \vspace{-0.1in}
\end{figure*}

%% file: 3_method.tex



\vspace{-0.1in}
\section{Method}
\vspace{-0.1in}
Similar to the original RPN, our method has two major components, see Figure~\ref{fig:main}. The first consists of a feature extractor that takes a grid of radiance and density sampled from NeRF as input, and produces a feature pyramid as output. The second is an RPN head which operates on the feature pyramid and generates object proposals. The feature volumes corresponding to the proposals can subsequently be extracted and processed for any downstream tasks. Our method is flexible in the form of NeRF input features, the feature extractor architectures, and the RPN module, which can be adapted to multiple downstream tasks.

\subsection{Input Sampling from NeRF}

Our method assumes a fully-trained NeRF model with reasonable quality model is provided.  The first step is to uniformly sample its radiance and density field to construct a feature volume. Despite different NeRF variants exist with different radiance field representations or structures, they share the same property that radiance and density can be queried with view directions and spatial locations. As essentially the radiance and density go through a similar volumetric rendering process, 
our method uses them as the input so that it is agnostic to concrete NeRF structures.

We uniformly sample the radiance and density on a 3D grid that covers the full traceable volume of the NeRF model, which is determined by slightly enlarging the bounding box containing all the cameras and objects in the scene. The resolution of the grid in each dimension is proportional to the length of the traceable volume in that dimension so that the aspect ratio of the objects are maintained. 
For NeRF models using plain RGB as radiance representation, we sample from the same set of viewing directions according to the camera poses used in NeRF training and average the results. If such camera poses are unknown, we uniformly sample directions from a sphere. Generally, each sample can be written as $(r, g, b, \alpha)$, where $(r, g, b)$ is the averaged radiance and $\alpha$ is converted from the density $\sigma$:
\begin{equation}
    \alpha = \clip{(1 - \exp(-\sigma\delta), 0, 1)},
\end{equation}
where $\delta = 0.01$ is a preset distance. For NeRF models adopting spherical harmonics or other basis functions as radiance representation, either the computed RGB values or the coefficients of the basis functions can be used as radiance information, depending on the downstream task.

\subsection{Feature Extractors}
Given the sampled grid, the feature extractor will generate a feature pyramid. We adopt three backbones: VGG\cite{simonyan2014vgg}, ResNet\cite{he2015Resnet} and Swin Transformer\cite{liu2021Swin} in our experiments, but other backbone networks may also be applicable. Considering the large variation in object sizes for indoor NeRF scenes as well as the scale differences between different NeRF scenes, we incorporate an FPN~\cite{Lin_2017_CVPR} structure to generate multi-scale feature volumes and to infuse high-level semantics into higher resolution feature volumes. For VGG, ResNet, and the FPN layers, we replace all the 2D convolutions, poolings, and normalization layers with their 3D counterparts. For Swin Transformer, we correspondingly employ 3D position embedding and shifted windows. 

\subsection{3D Region Proposal Networks}
Our 3D Region Proposal Network takes the feature pyramid from the feature extractor and outputs a set of oriented bounding boxes (OBB) with corresponding objectness scores. As in most 3D object detection works, we constrain the bounding box rotation to $z$-axis only (yaw angle), which is aligned with the gravity vector and perpendicular to the ground. We experiment two types of region proposal methods: anchor-based and anchor-free method, see Figure~\ref{fig:head_midpoint}.

\input{figs/fig-head+midpoint}

\vspace{2mm}
\noindent\textbf{Anchor-Based RPNs}
Conventional RPNs as  originally proposed in Faster R-CNN~\cite{ren2015faster} place anchors of different sizes and aspect-ratios at each pixel location and predict objectness scores and bounding box regression offsets for each anchor. We extend this approach to 3D by placing 3D anchors of different aspect-ratios and scales in voxels on different levels of the feature pyramid. We add $k$ levels of 3D convolutional layers after the feature pyramids (typically $k = 2$ or $4$), on top of which two separate $1 \times 1 \times 1$ 3D convolutional layers are used to predict the probability $p$ that an object exists, and the bounding box offsets $\boldsymbol t$ for each anchor, see Figure~\ref{fig:head_midpoint}(a). These layers are shared between different levels of the feature pyramid to reduce the number of parameters and improve the robustness to scale variation. The bounding box offsets $\boldsymbol t = (t_x, t_y, t_z, t_w, t_l, t_h, t_\alpha, t_\beta)$ are parametrized similarly as in \cite{xie2021oriented} but extended with a new dimension:
\begin{equation}
\begin{alignedat}{2}
    & t_x = (x - x_a) / w_a,\quad && t_y = (y - y_a) / l_a, \\
    & t_z = (z - z_a) / h_a, && t_w = \log(w / w_a), \\
    & t_l = \log(l / l_a), && t_h = \log(h / h_a), \\
    & t_\alpha = \Delta \alpha / w, && t_\beta = \Delta \beta / l,
\end{alignedat}
\end{equation}
where $x,y,w,l,\Delta\alpha,\Delta\beta$ describe the OBB projected onto the $xy$-plane, and $z, h$ are for the additional dimension in height. $x_a, y_a, z_a, w_a, l_a, h_a$ give the position and size of the reference anchor, see Figure~\ref{fig:head_midpoint}(a). Note that this encoding does not guarantee the decoded OBBs are cuboids. We follow~\cite{xie2021oriented} to transform the projections into rectangles before using them as proposals.

To determine the label of each anchor, we follow the process in Faster R-CNN but with parameters adapted under the 3D setting: we assign a positive label to an anchor if it has an Intersection-over-Union (IoU) overlap greater than 0.35 with any of the ground-truth boxes, or if it has the highest IoU overlap among all anchors with one of the ground-truth box. Non-positive anchors with IoU below 0.2 for all ground-truth boxes are regarded negative. Anchors that are neither positive nor negative are ignored in the loss computation. The loss is similar to that in Faster R-CNN:
\begin{align}
\begin{split}
    L(\{p_i\}, \{\boldsymbol t_i\}) &= \frac{1}{N_{cls}}\sum_i L_{cls}(p_i, p_i^*) \\ &+ \frac{\lambda}{N_{reg}}\sum_i p_i^* L_{reg}(\boldsymbol t_i, \boldsymbol t_i^*),
\end{split}
\label{eq:3}
\end{align}
where $p_i, \boldsymbol t_i$ are predicted objectness and box offsets, $p_i^*, \boldsymbol t_i^*$ are ground-truth targets, $N_{cls}, N_{reg}$ are the number of anchors involved in loss computation, and $\lambda$ is a balancing factor between the two losses. $L_{cls}$ is the binary cross entropy loss and $L_{reg}$ is the smooth $L_1$ loss in \cite{girshick2015fast}. The regression loss is only computed for positive anchors.


\vspace{2mm}
\noindent\textbf{Anchor-Free RPNs}
Anchor-free object detectors discard the expensive IoU computation between anchors and ground-truth boxes and can be used for region proposal in specific problem scopes (e.g., figure-ground segmentation).
We choose FCOS which is a representative anchor-free method and extend it to 3D.

Unlike anchor-based methods, our FCOS-based RPN predicts a single objectness $p$, a set of bounding box offsets $\boldsymbol t = (x_0, y_0, z_0, x_1, y_1, z_1, \Delta\alpha, \Delta\beta)$, and a centerness score $c$ for each voxel, see Figure~\ref{fig:head_midpoint}(b). We extend the encoding of box offsets in FCOS and define the regression target $\boldsymbol t^*_i = (x_0^*, y_0^*, z_0^*, x_1^*, y_1^*, z_1^*, \Delta\alpha^*, \Delta\beta^*)$ as following:
\begin{equation}
\begin{alignedat}{2}
    &x_0^* = x - x_0^{(i)},\quad && x_1^* = x_1^{(i)} - x, \\
    &y_0^* = y - y_0^{(i)},\quad && y_1^* = y_1^{(i)} - y, \\
    &z_0^* = z - z_0^{(i)},\quad && z_1^* = z_1^{(i)} - z, \\
    &\Delta \alpha^* = v_x^{(i)} - x,\quad && \Delta \beta^* = v_{y}^{(i)} - y,
\end{alignedat}
\end{equation}
where $x, y, z$ are the voxel position, $x_0^{(i)} < x_1^{(i)}$ are the left and right boundary of the \textbf{axis-aligned bounding box} (AABB) of $i$-th ground-truth OBB, and likewise for $y_0^{(i)}, y_1^{(i)}, z_0^{(i)}, z_1^{(i)}$. $v_x^{(i)}$ denotes the $x$ coordinate of the upmost vertex in the $xy$-plane projection of the OBB, and $v_y^{(i)}$ is the $y$ coordinate of the rightmost vertex, see Figure~\ref{fig:head_midpoint}(b). The ground-truth centerness is given by: 
\begin{equation}
\!\!\!\!    c^* = \sqrt{\frac{\min (x_0^*, x_1^*)}{\max (x_0^*, x_1^*)} \times \frac{\min (y_0^*, y_1^*)}{\max (y_0^*, y_1^*)} \times \frac{\min (z_0^*, z_1^*)}{\max (z_0^*, z_1^*)}}.
\end{equation}
The overall loss is then given by
\begin{align}
\begin{split}
    \!\!\!\!L(\{p_i\}, &\{\boldsymbol t_i\}, \{c_i\}) = \frac{1}{N_{pos}} L_{cls} (p_i, p_i^*) \\ &\!\!\!\!+ \frac{\lambda}{N_{pos}} p_i^* L_{reg} (\boldsymbol t_i, \boldsymbol t_i^*) + \frac{1}{N_{pos}} p_i^* L_{ctr} (c_i, c_i^*),
\end{split}
\label{eq:6}
\end{align}
where $L_{cls}$ is the focal loss in \cite{Lin_2017_ICCV} and $L_{reg}$ is the IoU loss for rotated boxes in \cite{zhou2019iou}. $L_{ctr}$ is the binary cross entropy loss. $p_i^* \in \{0, 1\}$ is the ground-truth label of each voxel in the feature pyramid, which is determined using the same center sampling and multi-level prediction process as in \cite{tian2019fcos}; $\lambda$ is the balancing factor and $N_{pos}$ is the number of voxels with $p_i^* = 1$. The regression and centerness loss only account for positive voxels.

To learn $p, \boldsymbol t, c$, we adapt the network in 
FCOS by adding $k = 2$ or $4$ 3D convolutional layers independently for the classification and regression branch after the feature pyramid. We append a convolutional layer on top of the classification and regression branch, respectively, to output $p$ and $\boldsymbol t$, and a parallel convolutional layer on the regression branch for predicting $c$. Like in our anchor-based method, we transform the possibly skewed box predictions into cuboids before further post-processing.

\subsection{Additional Loss Functions}

\input{figs/fig-objectness.tex}

\noindent\textbf{Objectness Classification} Although NeRF-RPN mainly targets on high recalls, some downstream tasks may prefer a low false-positive rate as well. To improve the precision of ROIs, we add a binary classification network as a sub-component to achieve foreground/background classification. More specifically, the network takes 1) the ROIs from RPN, and 2) the feature pyramid from the feature extractor as input, and outputs an objectness score and bounding box refinement offsets for each ROI, see Figure~\ref{fig:obj}. We extract rotation invariant features for each proposal via rotated ROI pooling. Each proposal is parameterized by $(x_r,y_r,z_r,w_r,l_r,h_r,\theta_r)$, where  $(x_r,y_r,z_r)$ describes the center coordinate, $w_r,l_r,h_r$ are the three dimensions, and $\theta_r \in [-\frac{\pi}{2}, \frac{\pi}{2})$ is the yaw angle.  
Referring to ~\cite{xie2021oriented}, we first enlarge the box and locate it in the corresponding feature volume, 
then apply trilinear interpolation to calculate the value on each feature point, and pad the ROI feature volume with zero before forwarding it to a pooling layer. The feature volume is pooled to $N \times 3\times 3 \times 3$ and used for further regression and classification. Referring to~\cite{ding2019learning}, the bounding box offset $\boldsymbol{g} = (g_x, g_y, g_z, g_w,g_l,g_h,g_\theta)$ is defined as
\begin{equation}
\begin{alignedat}{7}
& g_x = ((x-x_r)\cos \theta_r + (&y-y_r)\sin \theta_r ) / w_r\\
& g_y = ((y-y_r)\cos \theta_r - (&x-x_r)\sin \theta_r ) / l_r, \\
& g_z = (z - z_r) / h_r,  &g_w = \log(w / w_r), \\
& g_l = \log(l / l_r),  &g_h = \log(h / h_r), \\
& g_\theta = (\theta-\theta_r) / 2\pi &
\end{alignedat}
\end{equation}
The classification layer estimates the probability over 2 classes (namely, \textit{non-object}  and \textit{object} class). ROIs with IoU overlap greater than 0.25 with any of the ground-truth boxes are labeled as \textit{object}, while all the others are labeled \textit{non-object}. The loss function is similar to Equation~\ref{eq:3}, where the box offsets are replaced by $\boldsymbol{g}, \boldsymbol{g}^*$.

\vspace{2mm}
\noindent\textbf{2D Projection Loss} We project 3D bounding box coordinates $b_i=(x_{i},y_{i},z_{i})$ into 2D $b'_i=(x'_{i},y'_{i})$ 
and construct a 2D projection loss as following:
\begin{equation}
L_{2d\ proj}(\{b'_i\}) = \frac{1}{N_{cam}N_{box}} L_{reg} (b'_i, b^{\prime *}_i),
\end{equation}
where $N_{cam}, N_{box}$ are the number of cameras and the number of proposals. We set 4 cameras at 4 top corners of the room, pointing to the room center. Refer to the supplementary material for more discussion.


%% file: figs/fig-head+midpoint.tex
\begin{figure}[t]
    \subfloat[Anchor-based 3D RPN head and bounding box representation]{%
      \includegraphics[width=1\columnwidth]{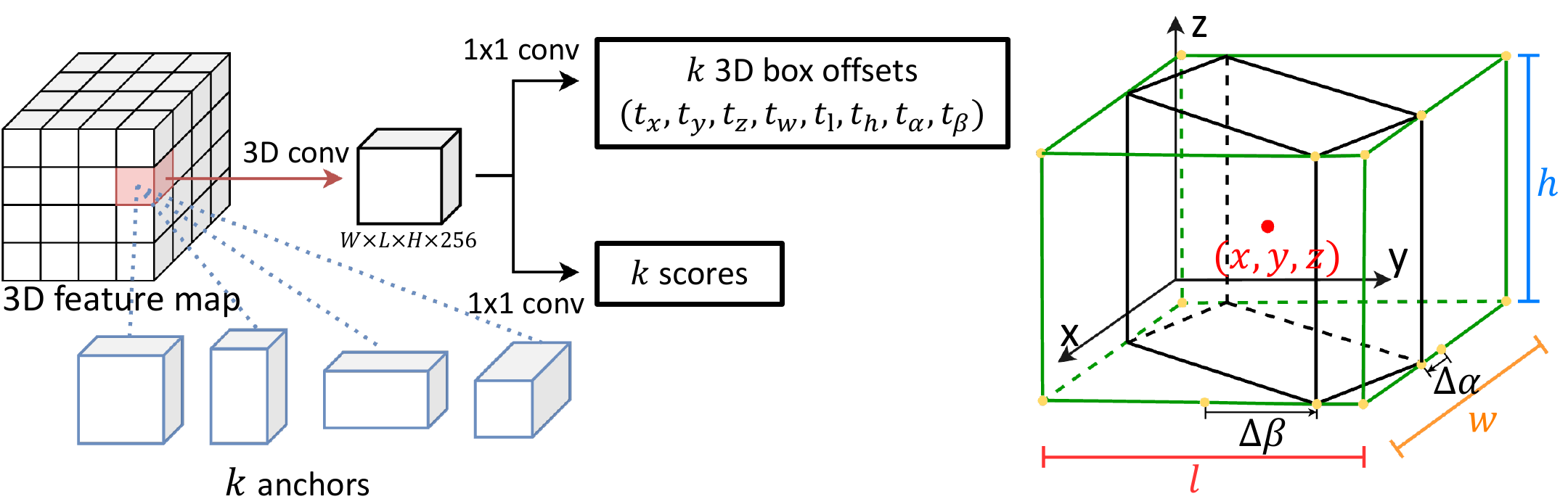}%
      \label{fig:head_midpint_a}
    }
    \vspace{0.1in}
    
    \subfloat[Anchor-free 3D RPN head and bounding box representation]{%
      \includegraphics[width=1\columnwidth]{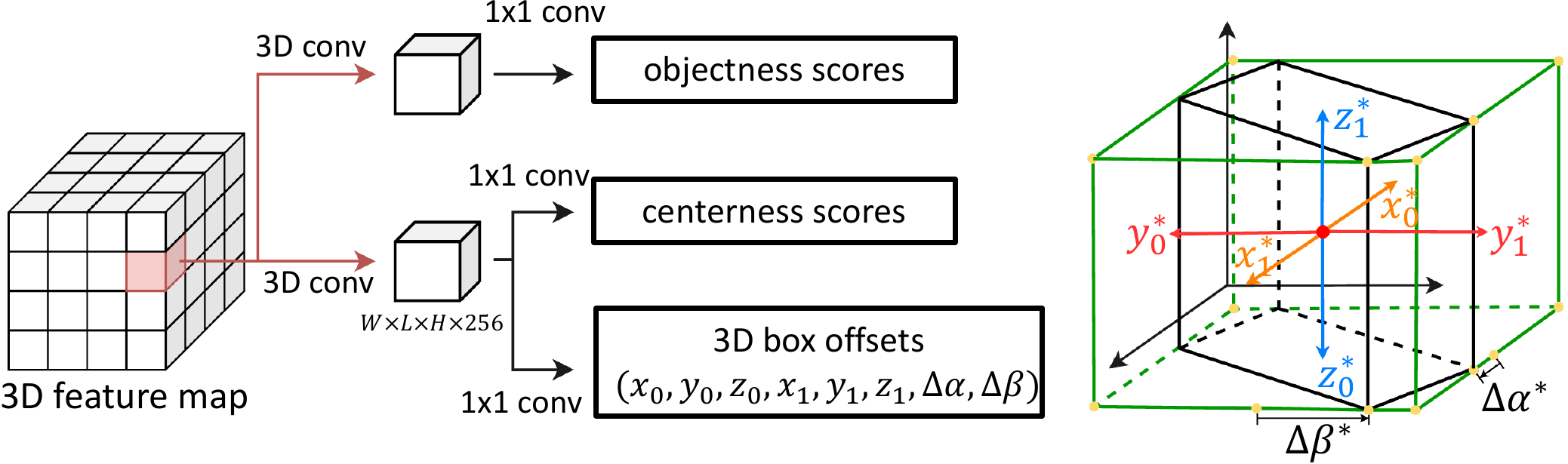}%
      \label{fig:head_midpint_b}
    }
    \vspace{-0.1in}
    \caption{\textbf{3D RPN Head.} These two figures illustrate the architectures of anchor-based and anchor-free 3D RPN heads along with their 3D midpoint offset bounding box representations.} \vspace{-0.15in}
    \label{fig:head_midpoint}
    
\end{figure}

%% file: figs/fig-objectness.tex
\begin{figure}[t]
    \includegraphics[width=1\linewidth]{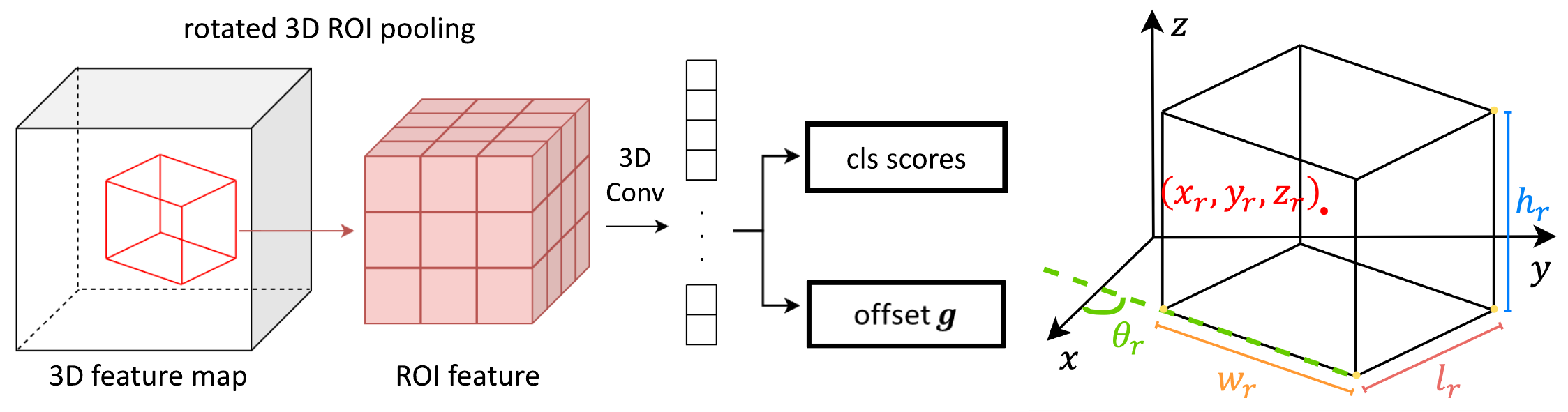}
    \vspace{-0.25in}
    \caption{\textbf{Binary Classification Network.} The binary classification network architecture with rotated 3D ROI pooling along with the bounding box representation used in this network. }
    \vspace{-0.2in}
    \label{fig:obj}
\end{figure}

%% file: 4_dataset.tex
\section{\nerf Dataset for 3D Object Detection}


There has been no  representative \nerf dataset constructed for 3D object detection. Thus, we build the first \nerf dataset for 3D object detection utilizing Hypersim~\cite{roberts:2021} and 3D-FRONT~\cite{fu20213d} datasets. In addition to these synthetic datasets, we  incorporate a subset of the real-world datasets from SceneNN~\cite{hua2016scenenn} and ScanNet~\cite{dai2017scannet} to demonstrate that our method is robust to real-world data. Figure~\ref{fig:dataset} shows some selected examples of the 3D groundtruth boxes we carefully labeled from 3D-FRONT.
Table~\ref{tab:dataset} summarizes our dataset. 
\input{tabs/tab-dataset.tex}
\input{figs/fig-dataset}

\noindent\textbf{Hypersim} Hypersim is a very realistic synthetic dataset for indoor scene understanding containing a wide variety of rendered objects with 3D semantics. However, the dataset is not specifically designed for NeRF training, where the object annotations provided are noisy for direct use in region proposal tasks. Thus, we perform extensive cleaning based on both the NeRF reconstruction quality and the usability of object annotations \ifshowsolution
\else
 (see supp mtrl).
\fi
Finally we keep around 250 scenes after cleanup.
The original 3D object bounding boxes in Hypersim are not carefully pruned, as some objects are invisible in all images. Furthermore, many instances are too fine in scale, while some are of less or little interest, e.g., floors and windows. We remove ambiguous objects that may interfere our training.
Then, we filter out tiny or thin objects by checking if the smallest dimension of their AABB is below a certain threshold. After these automatic pre-processing, we manually examine each remaining object. 
Objects that are visible in less than three images, or with over half of their AABBs invisible in all images, are 
removed. 

\vspace{2mm}
\noindent\textbf{3D-FRONT} 3D-FRONT \cite{fu20213d} is a large-scale synthetic indoor scene dataset with room layouts and textured furniture models. 
Due to its size, effort has been spent on splitting complex scenes into individual rooms and cleaning up bounding boxes
\ifshowsolution
\else
(supp mtrl).
\fi A total of 159 usable rooms are manually selected, cleaned, and rendered in our dataset. More rooms can be generated for \nerf training using our code and 3D-FRONT dataset, which will be released when the paper is accepted for publication.
We perform extensive manual cleaning on the  bounding boxes in each room. Similar to Hypersim, bounding boxes for construction objects such as ceilings and floors are removed automatically based on their labels. Moreover, 
we manually merge the relevant parts bounding boxes to label the entire semantic object (e.g., seat, back panel and legs are merged into a chair box). 
Refer to Figure~\ref{fig:dataset} for examples.

\vspace{2mm}
\noindent\textbf{Real-World Dataset} We construct our real-world \nerf dataset leveraging ScanNet \cite{dai2017scannet}, SceneNN \cite{hua2016scenenn}, and a dataset from INRIA~\cite{inriadataset}. ScanNet is a commonly used real-world dataset for indoor 3D object detection which contains over 1,500 scans. We randomly select 90 scenes and for each scene, we uniformly divide the video frames into 100 bins and select the sharpest frame in each bin based on the variance of Laplacian. 
We use the provided depth and a depth-guided NeRF~\cite{roessle2022depthpriorsnerf} to train the models. For object annotations, we compute the minimum bounding boxes based on the annotated meshes and discard objects of certain classes and sizes as similarly done for Hypersim.

%% file: tabs/tab-dataset.tex
\begin{table}[h]
\resizebox{0.99\linewidth}{!}{
\begin{tabular}{cccccccc}
\hline
\multirow{3}{*}{Datasets} & \multirow{3}{*}{\# Scenes} & \multicolumn{6}{c}{\# Boxes}                                                                                                                                                \\ \cline{3-8} 
                          &                            & \multirow{2}{*}{Total \#} & \multirow{2}{*}{\begin{tabular}[c]{@{}c@{}}Average \#\\ (per scene)\end{tabular}} & \multicolumn{4}{c}{\# Boxes in size (\# voxels)}            \\ \cline{5-8} 
                          &                            &                           &                                                                                   & $<16^3$ & $16^3$$\sim$$32^3$ & $32^3$$\sim$$64^3$ & $>64^3$ \\ \hline
Hypersim                  & 250                        & 4798                      & 19.2                                                                              & 3836    & 770                & 184                & 8       \\
3D-FRONT                  & 159                        & 1191                      & 7.5                                                                               & 129     & 703                & 324                & 35      \\
ScanNet                   & 90                         & 1086                      & 12.1                                                                              & 508     & 488                & 88                 & 2       \\
SceneNN                   & 16                         & 367                       & 22.9                                                                              & 182     & 112                & 54                 & 19      \\ \hline
\end{tabular}}
\vspace{-0.1in}
\caption{\textbf{Statistics of our \nerf dataset for 3D Object Detection.} }
\vspace{-0.1in}
\label{tab:dataset}
\vspace{-0.05in}
\end{table}

%% file: figs/fig-dataset.tex
\newcommand\x{1.8cm}
\newcommand\s{3.5cm}
\begin{figure}[t]
\centering
	\subfloat{\includegraphics[width = 0.49\linewidth]{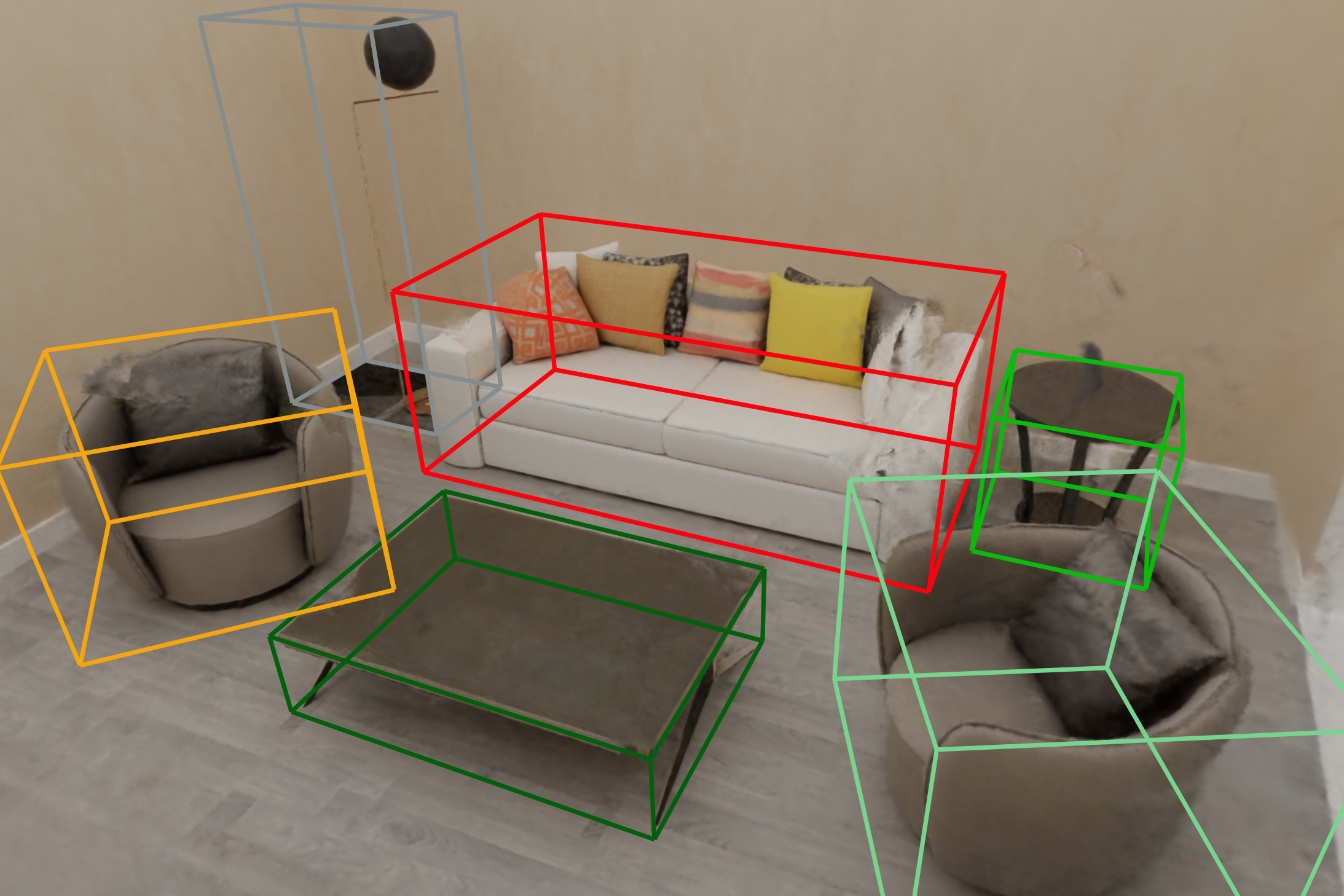}}
	\hfill
	\subfloat{\includegraphics[width = 0.49\linewidth]{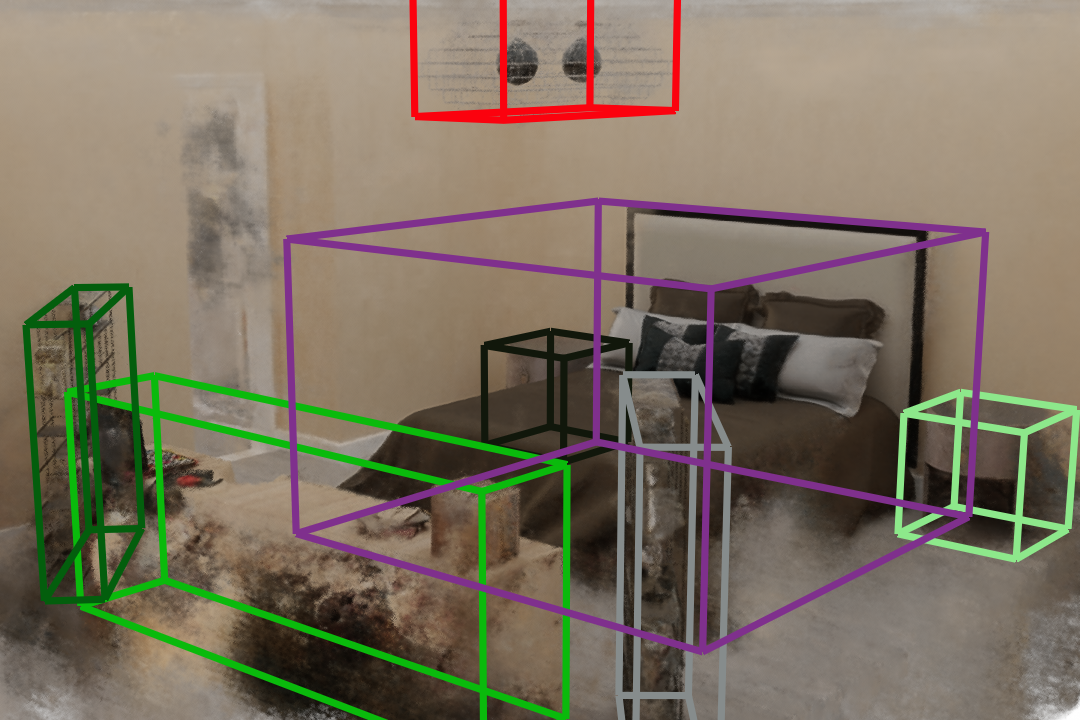}}
	
	\vspace{0.1cm}
	
	\subfloat{\includegraphics[width = 0.49\linewidth]{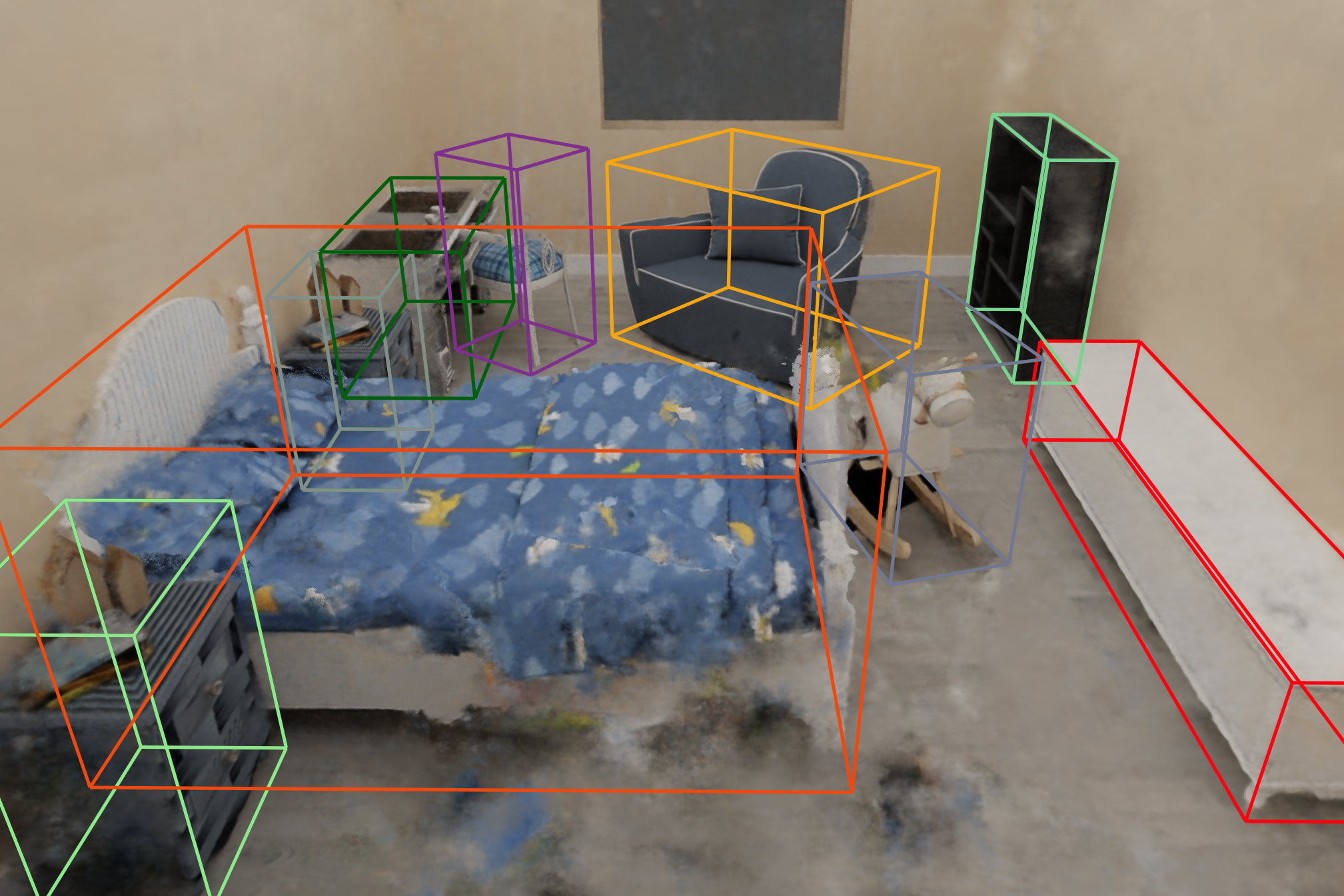}}
	\hfill
	\subfloat{\includegraphics[width = 0.49\linewidth]{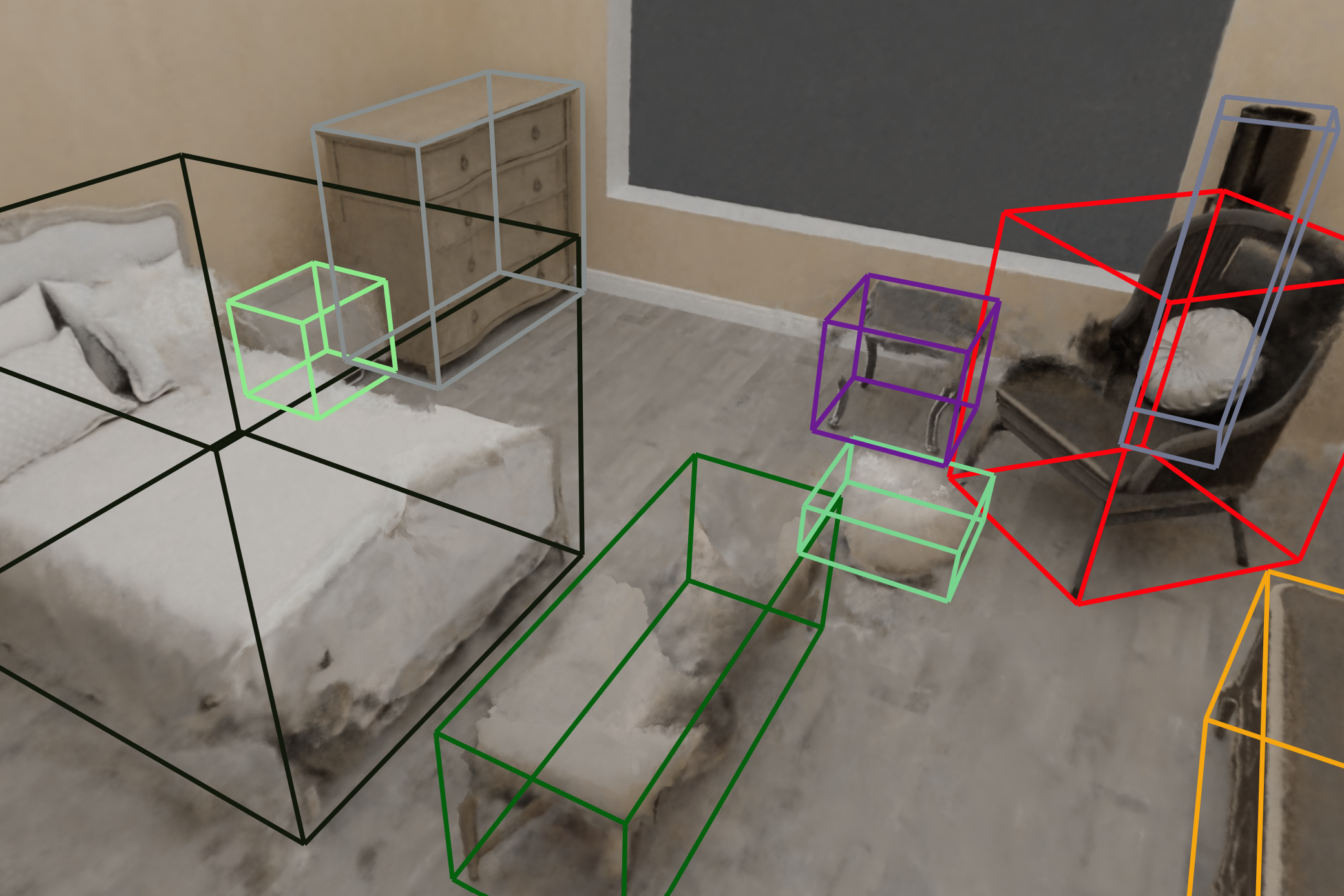}} 

	\vspace{0.2cm}

	\subfloat{\includegraphics[height = \x]{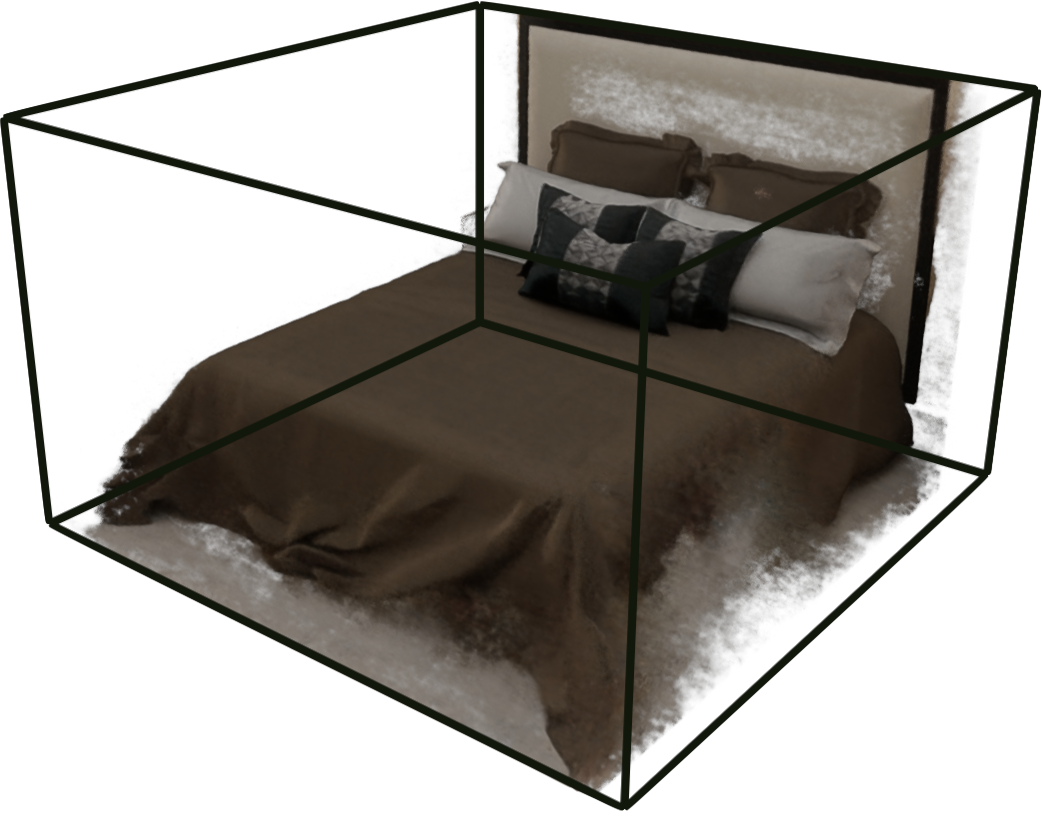}}
	\hfill
	\subfloat{\includegraphics[height = \x]{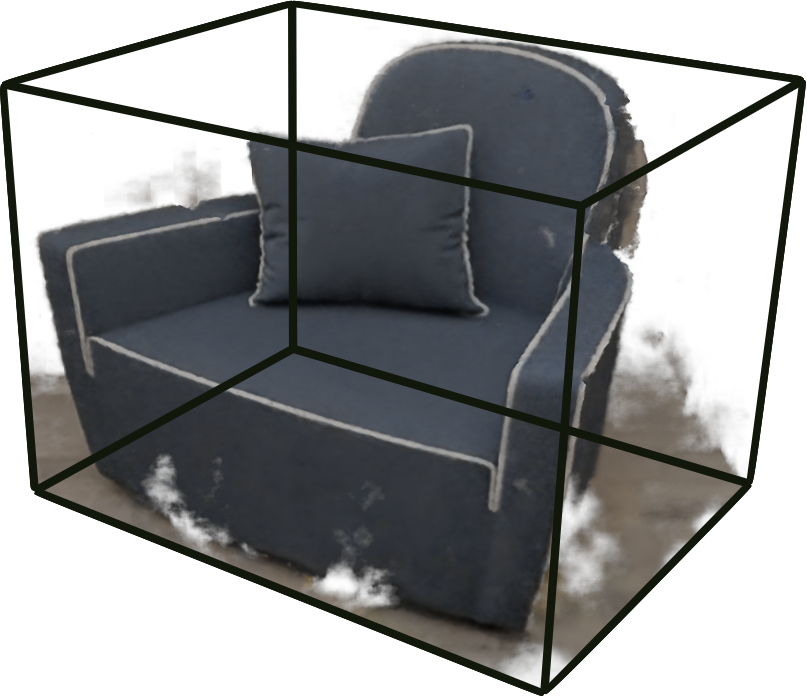}}
	\hfill
	\subfloat{\includegraphics[height = 1.7cm]{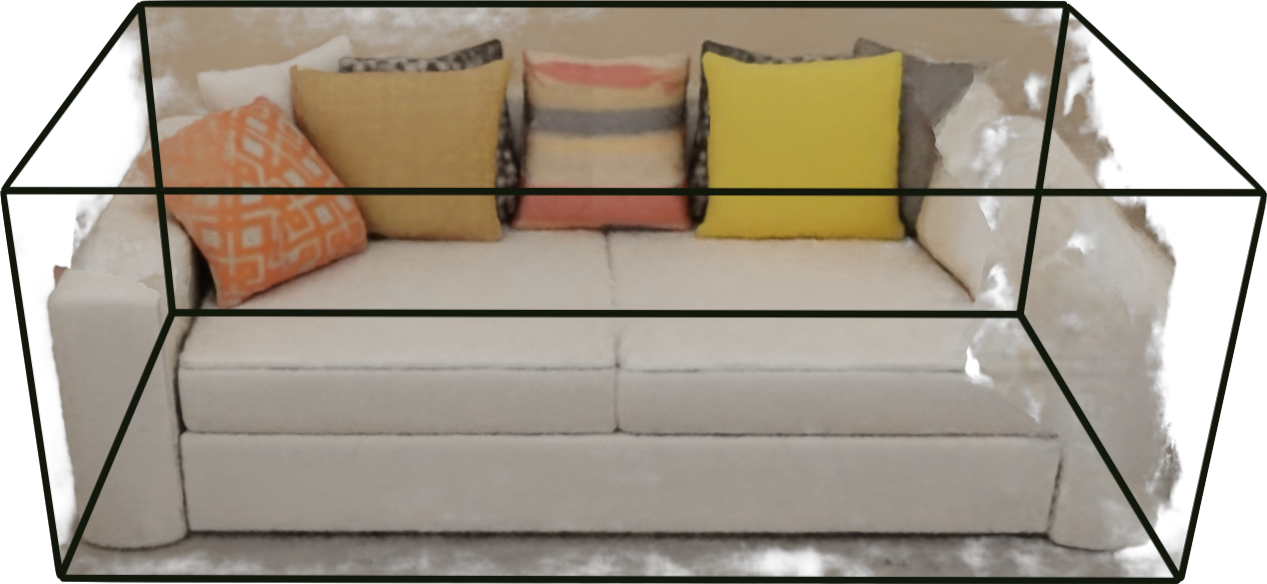}}
	
	\vspace{0.1cm}
	
	\subfloat{\includegraphics[height = 1.9cm]{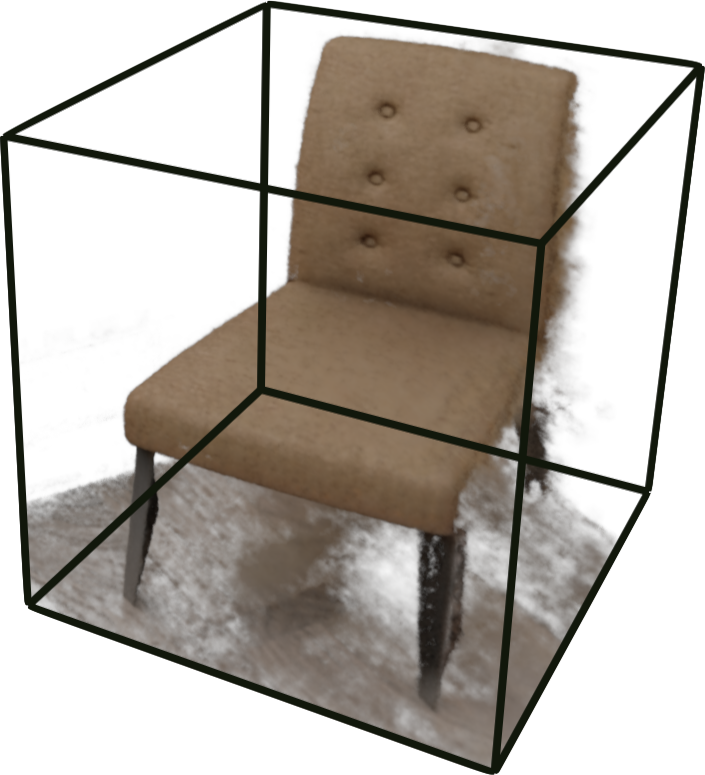}} 
	\hfill
	\subfloat{\includegraphics[height = 1.6cm]{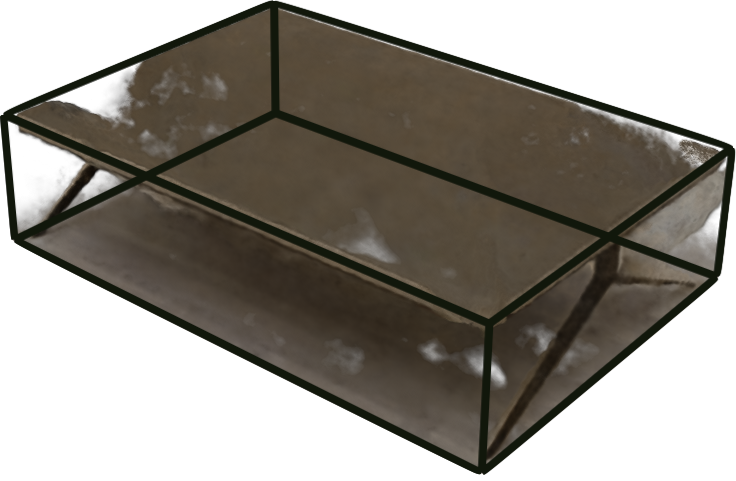}} 
	\hfill
	\subfloat{\includegraphics[height = 1.6cm]{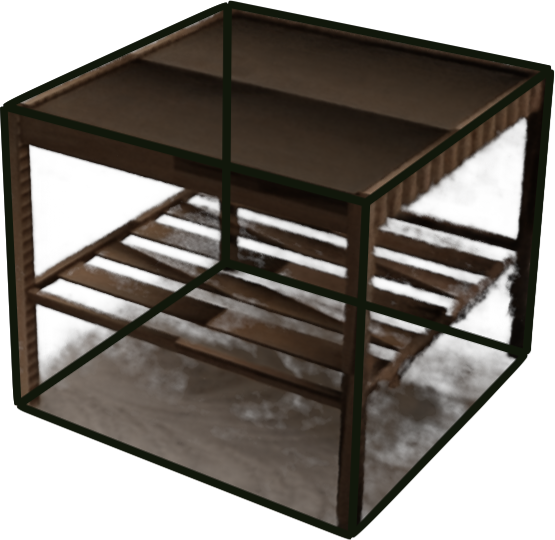}}
	\hfill
	\subfloat{\includegraphics[height = \x]{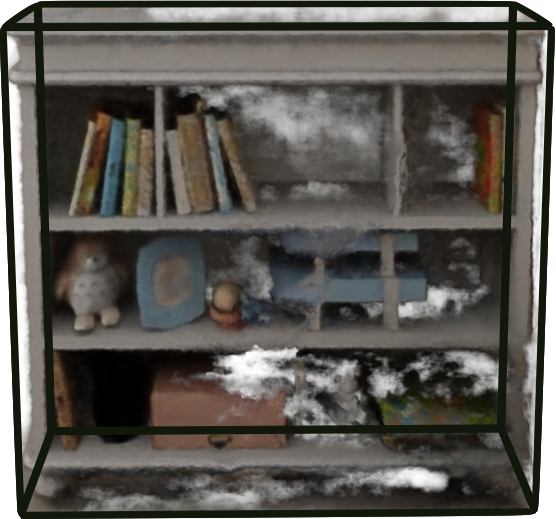}}
	
	\vspace{-0.1in}
    \caption{\textbf{3D-FRONT \nerf Dataset Samples. } Rows 1--2 show the \nerf reconstruction quality and ground-truth bounding box quality of our 3D-FRONT NeRF dataset. Rows 3--4 show groundtruth boxes with diverse object appearance in the dataset.}\vspace{-0.15in}
    \label{fig:dataset}
\end{figure}

%% file: 5_experiments.tex
\input{tabs/tab-compare_main}

\section{Experiments}


\subsection{Training \& Testing}

\noindent\textbf{Training} 
During training, input scenes are randomly flipped along $x,y$ axes and rotated along $z$-axis by $\frac{\pi}{2}$ with probability 0.5 for each augmentation operation. Additionally, the scenes are slightly rotated along $z$-axis by $\alpha \in [-\frac{\pi}{18}, \frac{\pi}{18}]$ with a probability of 0.5, which we find can 
significantly improve the average precision (AP) in RPN outputs.  We optimize our network with AdamW \cite{LoshchilovH19adamw} with an initial learning rate of 0.0003 and a weight decay of 0.001. In our training, we set $\lambda = 5.0$ for Eq.~(\ref{eq:3}) and $\lambda = 1.0$ for Eq.~(\ref{eq:6}). For the anchor-based approach, we adopt a 4-level FPN and anchors of 13 different aspect ratios, which are 1:1:1, 1:1:2, 1:1:3, 2:2:1, 3:3:1, and their permutations. All anchors on the same level of feature volume share the same size for their shortest side, which is in $\{8, 16, 32, 64\}$, from fine to coarse scale. Following the RPN training strategy in~\cite{ren2015faster}, we randomly sample 256 anchors from each scene in each iteration to compute the loss, where the ratio of positive and negative anchors is 1:1. For anchor-free approach, all output proposals are used to compute the loss.

\vspace{2mm}
\noindent\textbf{Testing}
After obtaining the ROIs with objectness scores, we first discard the boxes whose geometry centers are beyond the scene boundary. Then, we select the top 2,500 proposals on each level of the feature volumes independently. To remove redundant proposals, we apply Non-Maximum Suppression (NMS) to the aggregated boxes based on rotated-IoU with threshold 0.1, after which we select the 2,500 boxes with the highest objectness scores.

\input{figs/fig-qualitative.tex}
\input{figs/fig-failure.tex}

\subsection{Ablation Study}

\noindent\textbf{Backbones and Heads} Table~\ref{tab:ablation_archs} tabulates the recall and average precision of different combinations of feature extraction backbones and RPN heads. When fixing the backbones and comparing the RPN heads only, we observe that anchor-free models achieve a higher AP on all three datasets. The two RPN methods attain similar recalls on 3D-FRONT and ScanNet, while on Hypersim anchor-free models are generally higher in recalls. We believe the better performance of anchor-free models results are twofold: 1) The centerness prediction of anchor-free models helps suppress proposals that are off from the centers, which is particularly helpful when the bounding box center is misaligned with the mass center, or when the NeRF input is noisy; 2) The limited number of aspect ratios and scales for anchors limits the performance of anchor-based models as 3D objects vary greatly in sizes.

Furthermore, when comparing the performance between different backbones, we notice that models with VGG19 generally achieve better recall and AP compared to others. The major exception concerns the performance of anchor-based models on Hypersim, where Swin-S demonstrates superior recall and AP. Given that the NeRF results on Hypersim are significantly noiser and the scenes are more complex, we suspect that the larger receptive fields and the richer semantics enabled by the shifted windows, and attention of Swin Transformers are crucial to our anchor-based method in this case.

\vspace{1mm}
\noindent\textbf{NeRF Sampling Strategies}
While the density field from NeRF is view-independent, the radiance depends on the viewing direction and can be encoded with different schemes. In the supplemental material we investigate the effect of this view-dependent information, and conclude that using density alone is the best strategy.  

\input{tabs/tab-external-comparison}

\vspace{1mm}
\noindent\textbf{Regression Loss}
\input{tabs/tab-reg_loss}
We test three common loss functions for bounding box regression on the 3D-FRONT dataset using Swin-S as the backbone in Table~\ref{tab:reg_loss}. IoU loss directly optimizes the IoU between the predicted and ground-truth bounding boxes while DIoU loss \cite{zheng2020distance} penalizes the normalized distance between the two for faster convergence. We use the variants for oriented boxes of these two losses as proposed in~\cite{zhou2019iou}. Our results illustrate that IoU loss consistently outperforms the other two for the anchor-based approach, while for anchor-free models IoU and DIoU loss produce similar performance.

\vspace{1mm}
\noindent\textbf{Additional Loss}
We discuss the impact of aforementioned losses in the supplemental material.

\subsection{Results}
We performed experiments with different model configurations on various \nerf datasets constructed from Hypersim\cite{roberts:2021}, 3D-FRONT\cite{fu20213d}, ScanNet\cite{dai2017scannet}, SceneNN\cite{hua2016scenenn} and INRIA\cite{inriadataset}, where \cite{hua2016scenenn} and \cite{inriadataset} are only used in test time due to their relatively small numbers of usable scenes. Detailed quantitative results are shown in Table~\ref{tab:ablation_archs}. Figure~\ref{fig:qualitative} shows the qualitative results produced by the model with VGG19 backbone and anchor-free RPN head.
Figure~\ref{fig:failure} shows typical failure cases. During our experiments, we found that bad \nerf reconstruction can severely hamper the prediction. As aforementioned, the region proposal task largely depends on 3D geometry in \nerf. Similar to 2D RPN for images, our method also has missing/merging proposals or wrong rotation after NMS. 
Presently, our dataset handles first-level objects; tiny or second-level objects are future work.


\vspace{1mm}
\noindent\textbf{Application: Scene Editing} We can edit the scene in NeRF given the proposals produced by our NeRF-RPN. See Figure~\ref{fig:edit} for a demonstration which sets the density inside the proposal to zero when rendering.
\input{figs/fig-scene_editing.tex}

\subsection{Comparison}
To demonstrate the effectiveness of our method, we compare NeRF-RPN to recent 
3D object detection methods like ImVoxelNet \cite{rukhovich2022imvoxelnet} and FCAF3D \cite{rukhovich2022fcaf3d}, with results shown in Table~\ref{tab:external-comparison} and Figure~\ref{fig:external-comparison}. ImVoxelNet takes multi-view RGB as input, hence NeRF input images are used for training. FCAF3D is a point cloud-based 3D detector, so we use the ground-truth depth from Hypersim and ScanNet, NeRF rendered depth for 3D-FRONT, and the corresponding RGB images to build point clouds. We adapt the implementation of these two methods in~\cite{mmdet3d2020} and train them from scratch on the three datasets we use. Our method outperforms ImVoxelNet by a large margin on all datasets except 3D-FRONT, although ImVoxelNet takes advantage from our object-centric camera trajectory in 3D-FRONT. NeRF-RPN also outperforms FCAF3D on 3D-FRONT and Hypersim, despite ground-truth depth is used for Hypersim, giving FCAF3D extra advantages.

\input{figs/fig-external-comparison}

%% file: tabs/tab-compare_main.tex



\begin{table*}[]
\centering
\resizebox{0.99\linewidth}{!}{
\begin{tabular}{llllllllllllll}
\hline
\multirow{2}{*}{Methods} & \multirow{2}{*}{Backbones} & \multicolumn{4}{c}{Hypersim} & \multicolumn{4}{c}{3D-FRONT} & \multicolumn{4}{c}{ScanNet} \\
 &  & $\text{Recall}_{25}$ & $\text{Recall}_{50}$ & $\text{AP}_{25}$ & $\text{AP}_{50}$ & $\text{Recall}_{25}$ & $\text{Recall}_{50}$ & $\text{AP}_{25}$ & $\text{AP}_{50}$ & $\text{Recall}_{25}$ & $\text{Recall}_{50}$ & $\text{AP}_{25}$ & $\text{AP}_{50}$ \\ \hline
\multirow{3}{*}{Anchor-based} & VGG19 & 57.1 & 14.9 & 11.2 & 1.3 & 97.8 & \textbf{76.5} & 65.9 & 43.2 & 88.7 & 42.4 & 40.7 & 14.4 \\
 & ResNet-50 & 49.8 & 13.0 & 9.7 & 1.3 & 96.3 & 70.6 & 65.7 & 45.1 & 86.2 & 32.0 & 34.4 & 9.0 \\
 & Swin-S & 69.8 & \textbf{28.3} & 24.6 & 6.2 & \textbf{98.5} & 63.2 & 51.8 & 26.6 & \textbf{93.6} & \textbf{44.3} & 38.7 & 12.9 \\ \hline
\multirow{3}{*}{Anchor-free} & VGG19 & 66.7 & 27.3 & \textbf{30.9} & \textbf{11.5} & 96.3 & 69.9 & \textbf{85.2} & \textbf{59.9} & 89.2 & 42.9 & 55.5 & 18.4 \\
 & ResNet-50 & 63.2 & 17.5 & 23.2 & 6.0 & 95.6 & 67.7 & 83.9 & 55.6 & 91.6 & 35.5 & 55.7 & 16.1 \\
 & Swin-S & \textbf{70.8} & 21.0 & 27.7 & 7.7 & 96.3 & 62.5 & 78.7 & 41.0 & 90.6 & 39.9 & \textbf{57.5} & \textbf{20.5} \\ \hline
\end{tabular}}
\vspace{-0.1in}
\caption{\textbf{Ablation on different backbones and heads.} $\text{Recall}_{25}$ and $\text{Recall}_{50}$ denote the recall scores at an  IoU threshold of 0.25 and 0.5, respectively.}
\vspace{-0.0in}
\label{tab:ablation_archs}
\end{table*}

%% file: figs/fig-qualitative.tex
\newcommand\width{0.02cm}
\newcommand\height{0.06cm}
\newcommand{\nerfinput}{Input NeRF}
\newcommand{\heatmap}{Heat Map}
\newcommand{\proposals}{Proposals}

\begin{figure*}[ht]
    \centering
    \captionsetup[subfloat]{labelformat=empty}

    \subfloat{\includegraphics[width = 0.16\linewidth]{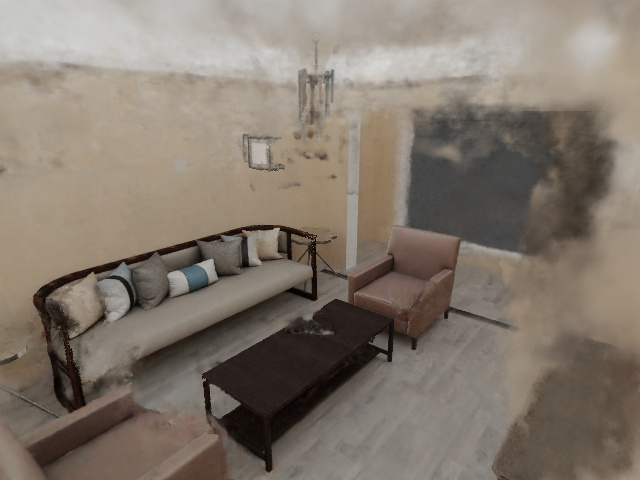}}\hspace{\width}
	\subfloat{\includegraphics[width = 0.16\linewidth]{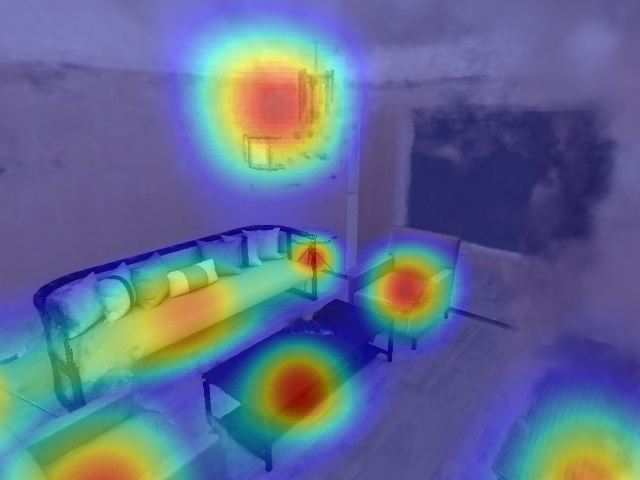}}\hspace{\width}
    \subfloat{\includegraphics[width = 0.16\linewidth]{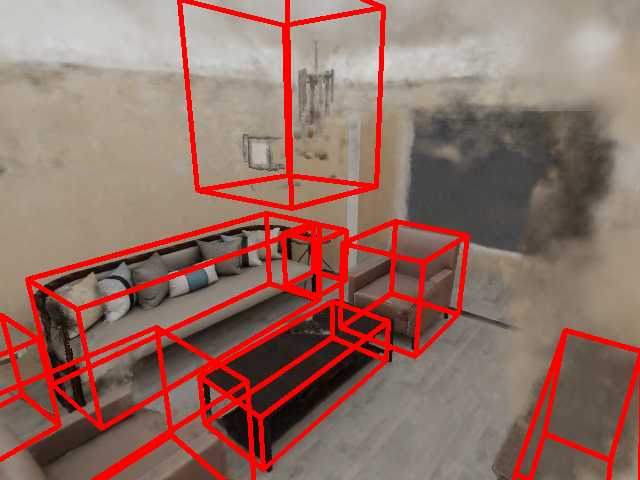}}
	\hfill
    \subfloat{\includegraphics[width = 0.16\linewidth]{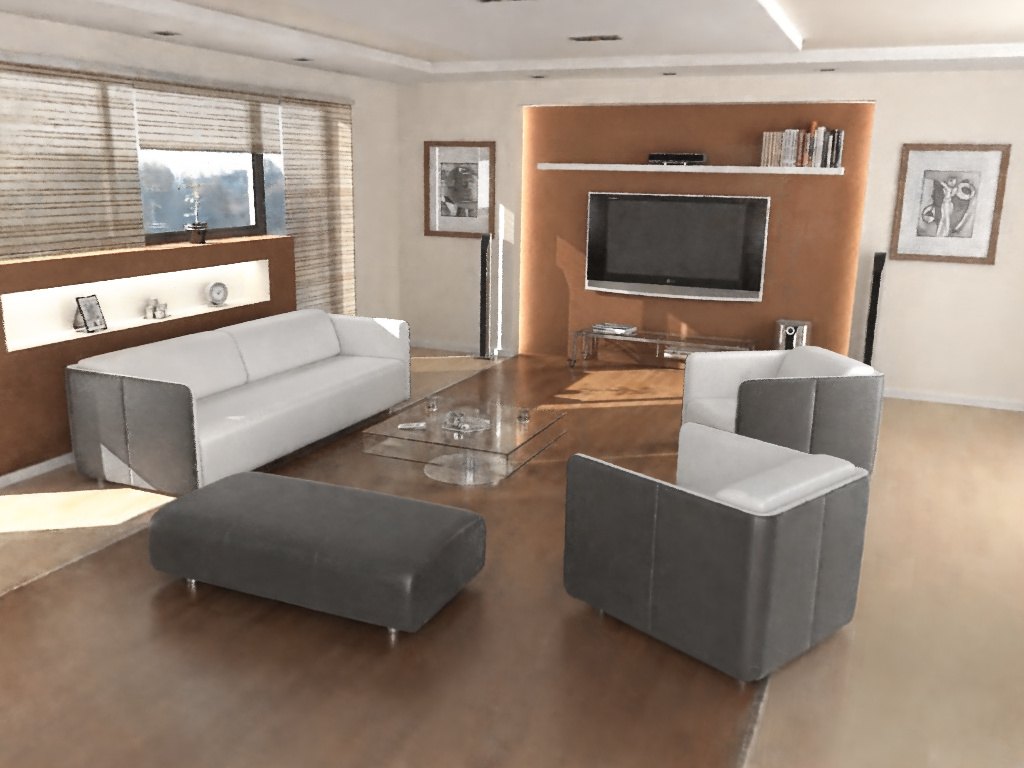}}\hspace{\width}
	\subfloat{\includegraphics[width = 0.16\linewidth]{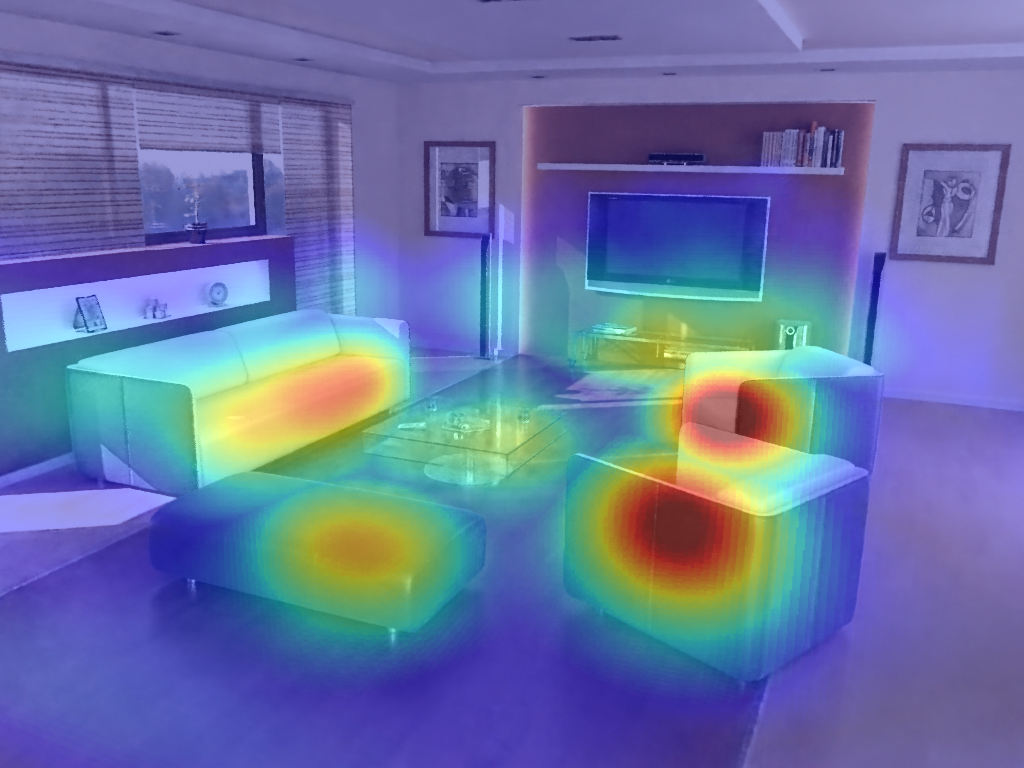}}\hspace{\width}
    \subfloat{\includegraphics[width = 0.16\linewidth]{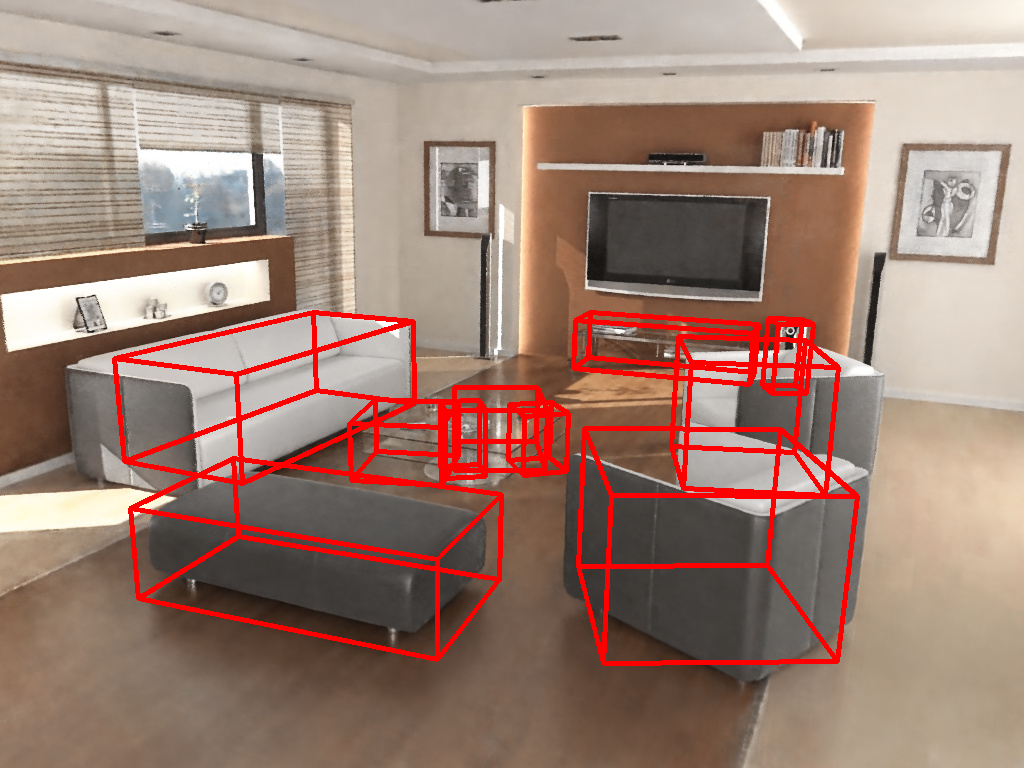}}
	
	\vspace{\height}

	\subfloat{\includegraphics[width = 0.16\linewidth]{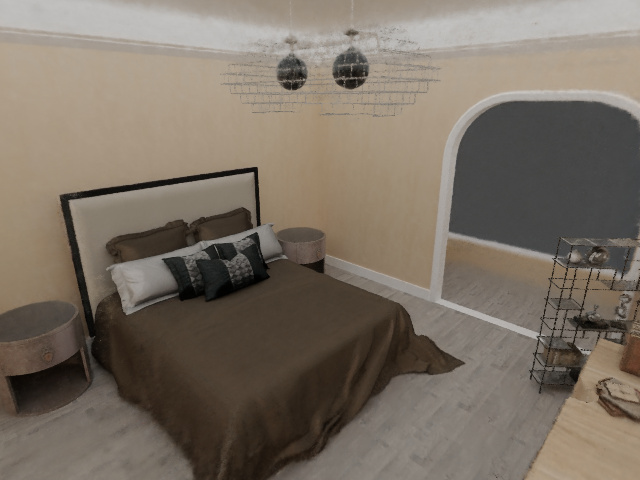}}\hspace{\width}
	\subfloat{\includegraphics[width = 0.16\linewidth]{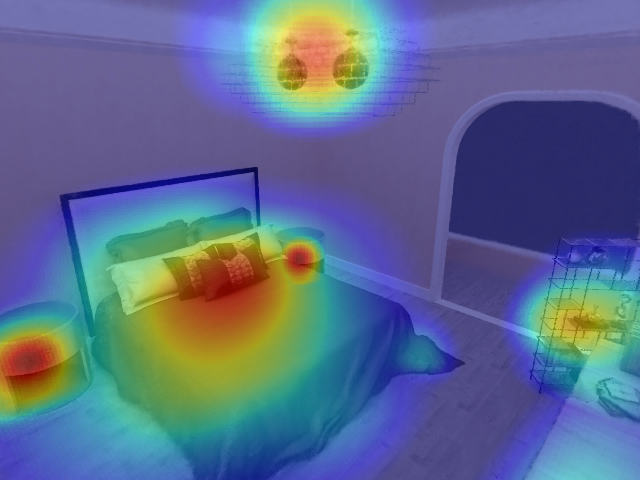}}\hspace{\width}
    \subfloat{\includegraphics[width = 0.16\linewidth]{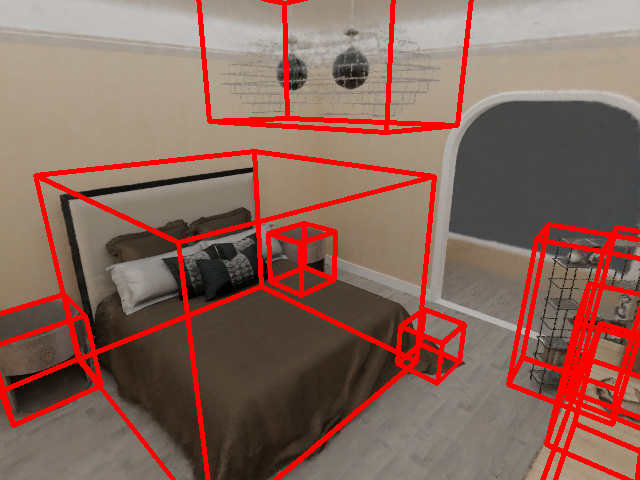}}
	\hfill
    \subfloat{\includegraphics[width = 0.16\linewidth]{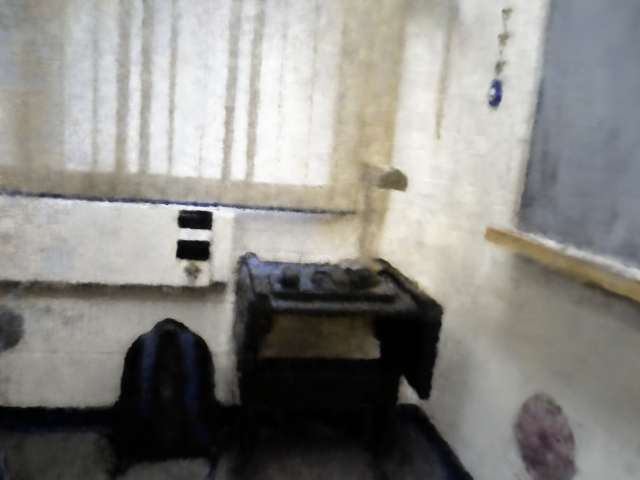}}\hspace{\width}
	\subfloat{\includegraphics[width = 0.16\linewidth]{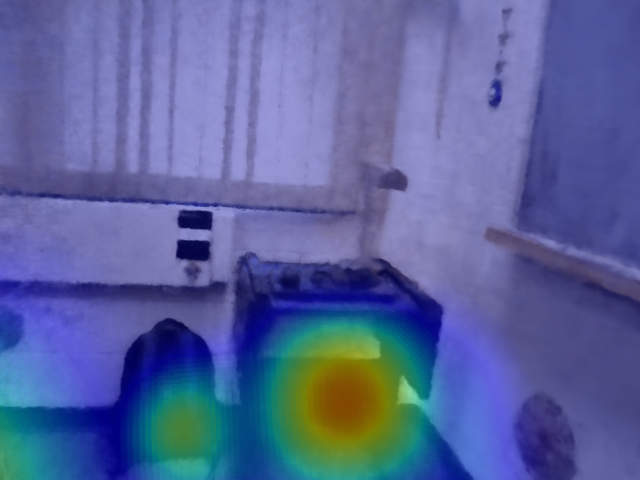}}\hspace{\width}
    \subfloat{\includegraphics[width = 0.16\linewidth]{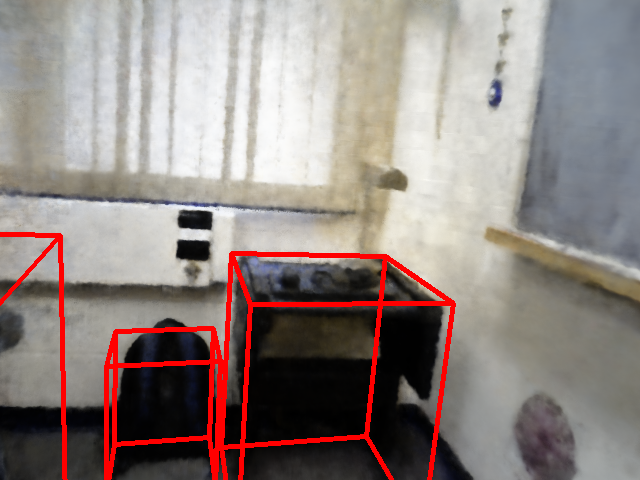}}
	
	\vspace{\height}

    \subfloat{\includegraphics[width = 0.16\linewidth]{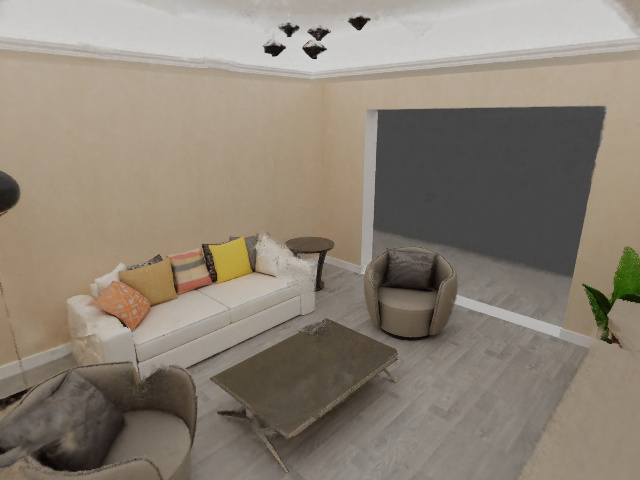}}\hspace{\width}
	\subfloat{\includegraphics[width = 0.16\linewidth]{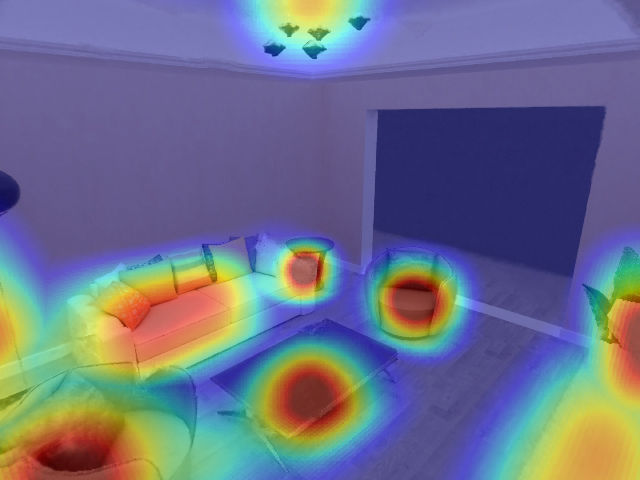}}\hspace{\width}
    \subfloat{\includegraphics[width = 0.16\linewidth]{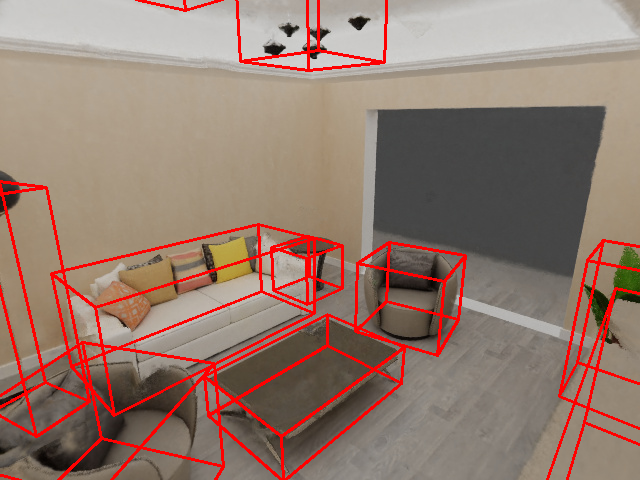}}
	\hfill
    \subfloat{\includegraphics[width = 0.16\linewidth]{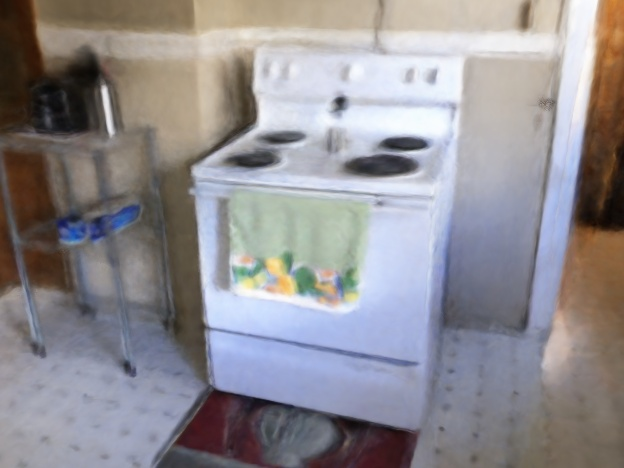}}\hspace{\width}
	\subfloat{\includegraphics[width = 0.16\linewidth]{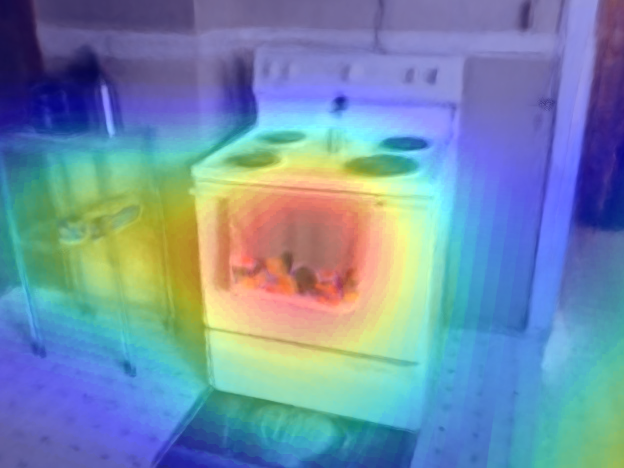}}\hspace{\width}
    \subfloat{\includegraphics[width = 0.16\linewidth]{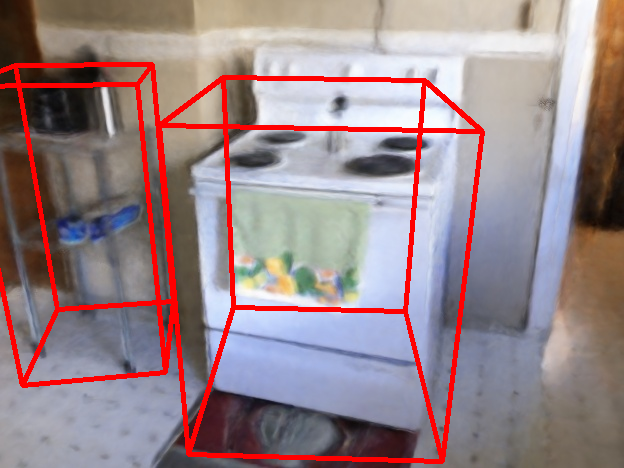}}
	
	\vspace{\height}
 
	\subfloat[\nerfinput]{\includegraphics[width = 0.16\linewidth]{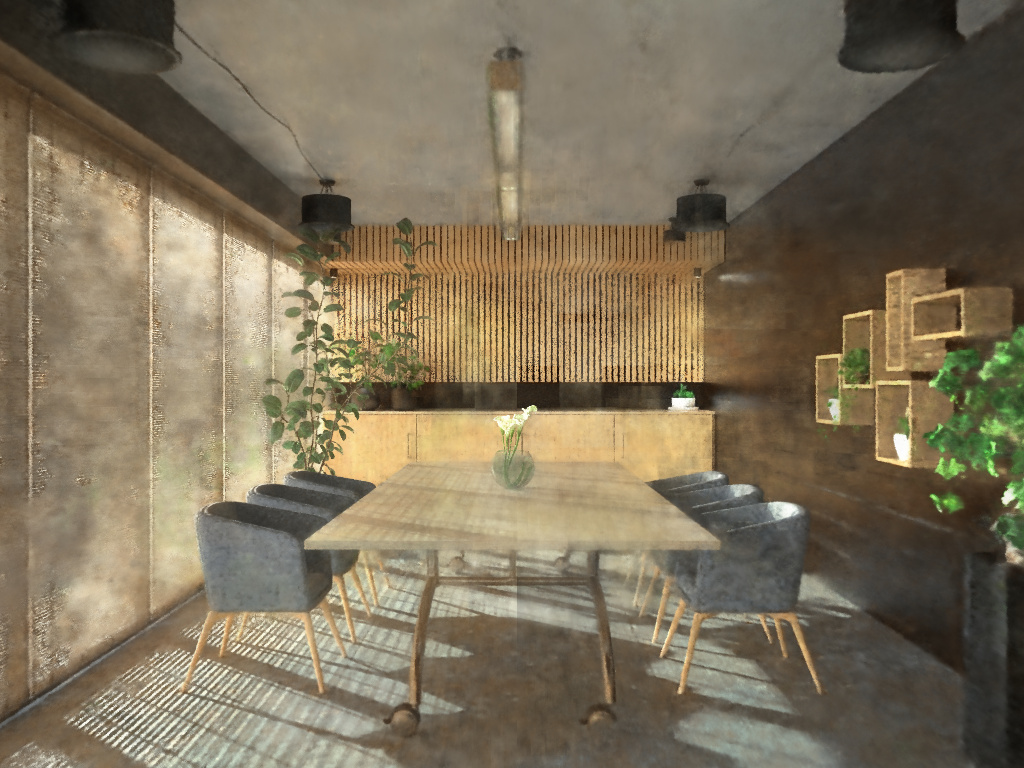}}\hspace{\width}
	\subfloat[\heatmap]{\includegraphics[width = 0.16\linewidth]{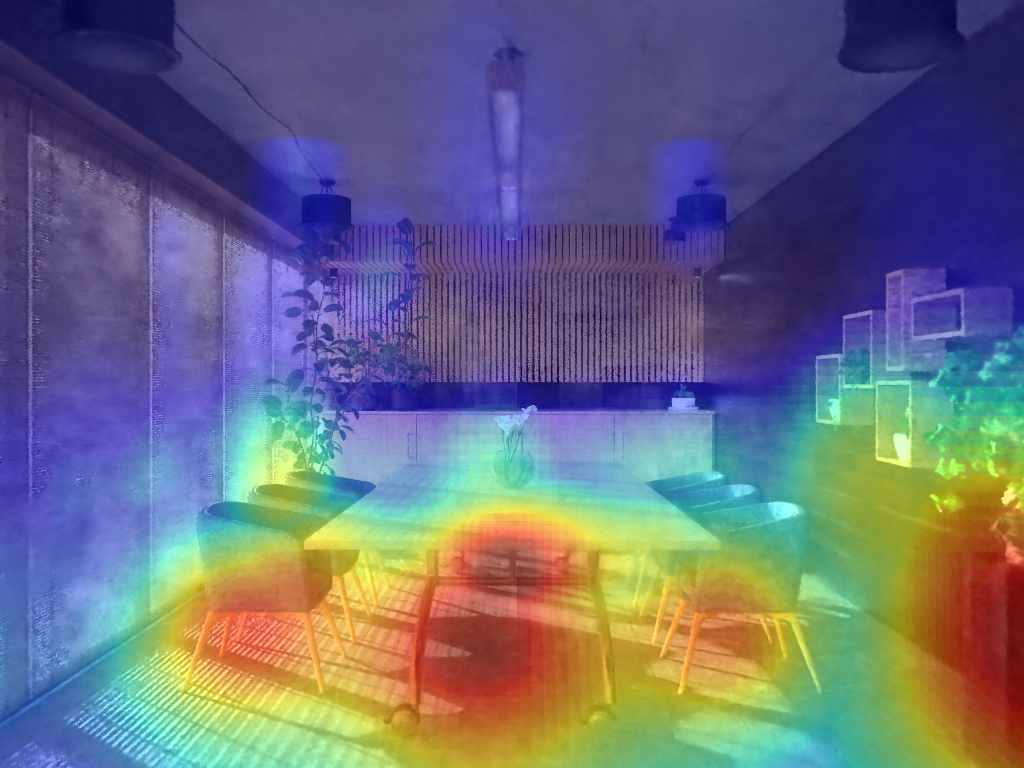}}\hspace{\width}
    \subfloat[\proposals]{\includegraphics[width = 0.16\linewidth]{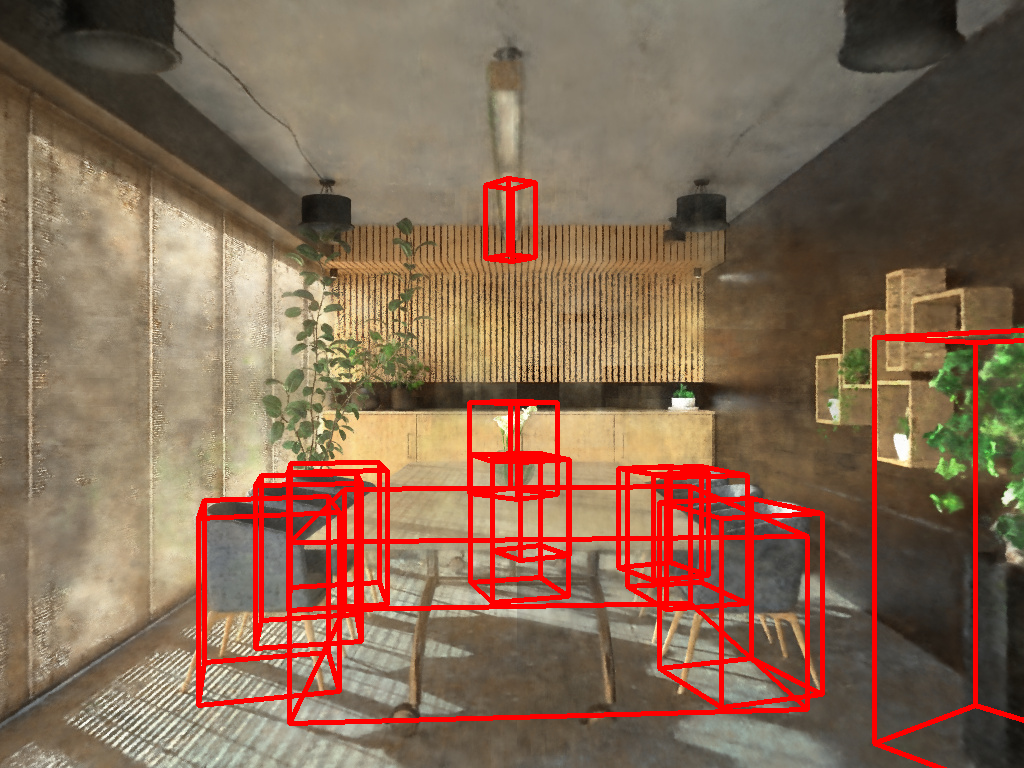}}
	\hfill
    \subfloat[\nerfinput]{\includegraphics[width = 0.16\linewidth]{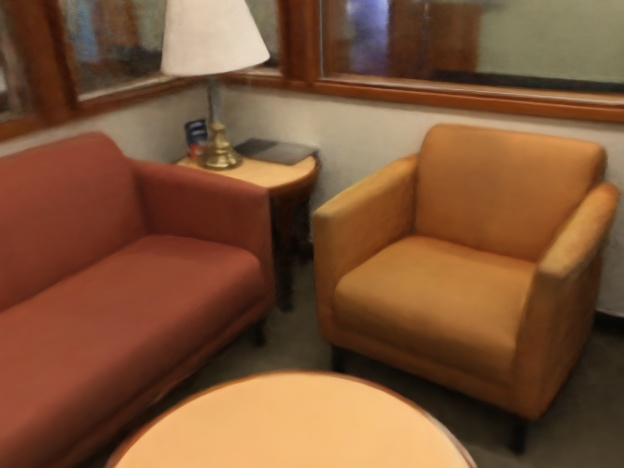}}\hspace{\width}
	\subfloat[\heatmap]{\includegraphics[width = 0.16\linewidth]{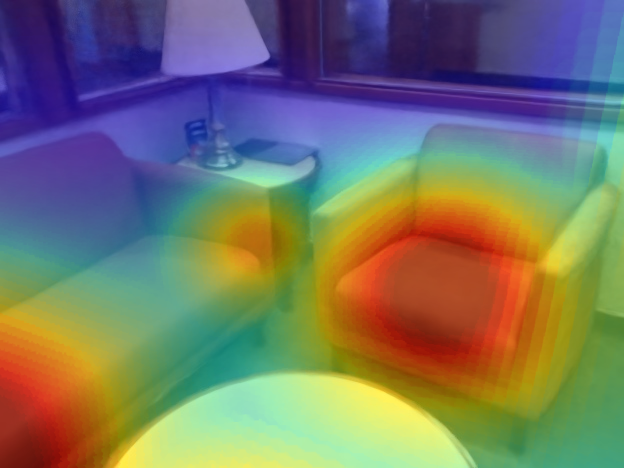}}\hspace{\width}
    \subfloat[\proposals]{\includegraphics[width = 0.16\linewidth]{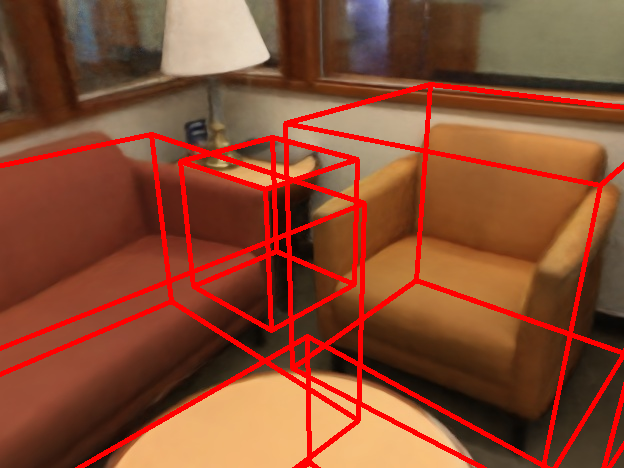}}
	
    \vspace{-0.1in}
    \caption{\textbf{Qualitative Results.} The ``Heat Map" columns show the distribution of proposal confidence scores where red means higher confidence. The ``Proposals" columns show a few top bounding boxes after NMS. From top to bottom, left to right, scene 1-3 are from 3D-FRONT, 4-5 from Hypersim, and 6-8 from ScanNet.}
    \vspace{-0.1in}
    \label{fig:qualitative}
\end{figure*}

%% file: figs/fig-failure.tex
\begin{figure}[ht]
    \centering
    
    \subfloat[]{\includegraphics[width = 0.24\linewidth]{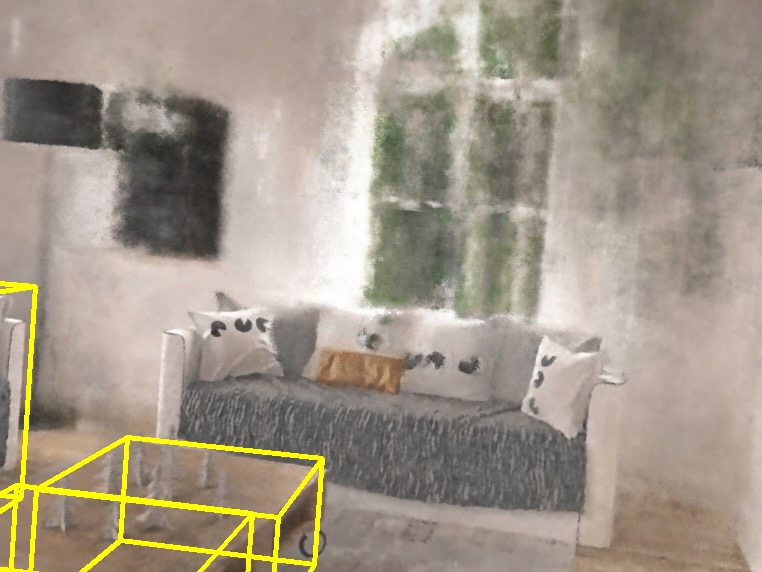}}\hfill
    \subfloat[]{\includegraphics[width = 0.24\linewidth]{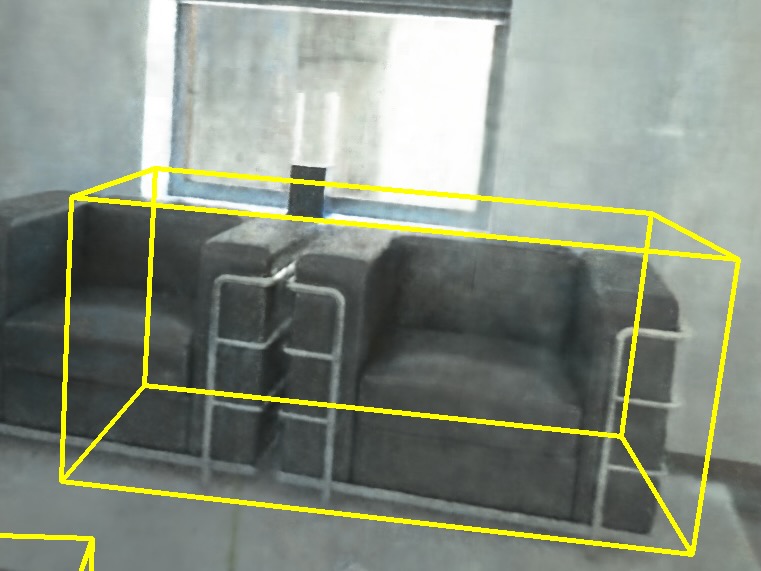}}\hfill
    \subfloat[]{\includegraphics[width = 0.24\linewidth]{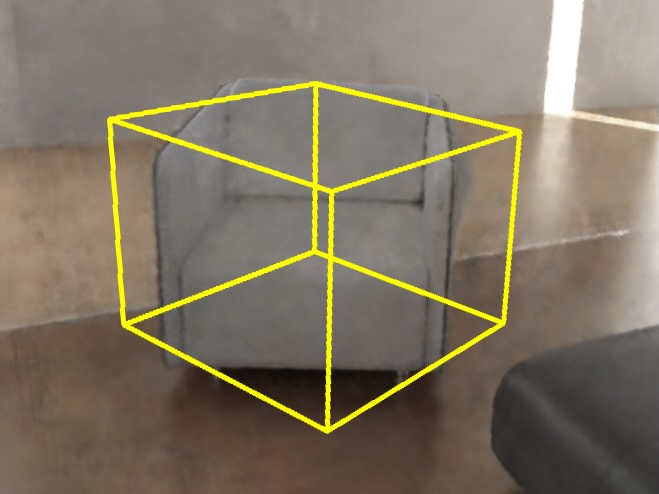}}\hfill
    \subfloat[]{\includegraphics[width = 0.24\linewidth]{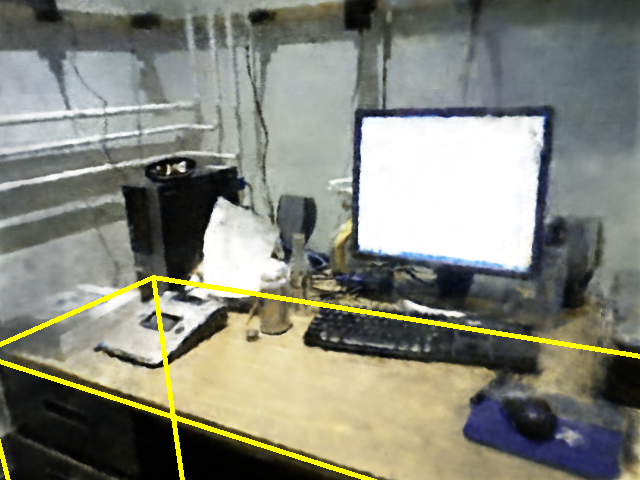}}\hfill
	
	\vspace{-0.1in}
    \caption{\textbf{Failure Cases.} (a)(b) Missing and merging proposals, (c) wrong rotation, (d) no proposal for tiny/second-level objects.}
    \vspace{-0.10in}
    \label{fig:failure}
\end{figure}

%% file: tabs/tab-external-comparison.tex
\begin{table*}[ht]
\centering
\resizebox{0.88\linewidth}{!}{
\begin{tabular}{lllllllllllll}
\hline
\multirow{2}{*}{Methods} & \multicolumn{4}{c}{Hypersim} & \multicolumn{4}{c}{3D-FRONT} & \multicolumn{4}{c}{ScanNet} \\
 & AR$_{25}$ & AR$_{50}$ & AP$_{25}$ & AP$_{50}$ & AR$_{25}$ & AR$_{50}$ & AP$_{25}$ & AP$_{50}$ & AR$_{25}$ & AR$_{50}$ & AP$_{25}$ & AP$_{50}$ \\ \hline
ImVoxelNet & 19.7 & 5.7 & 9.7 & 2.3 & 88.3 & 71.5 & \textbf{86.1} & \textbf{66.4} & 51.7 & 20.2 & 37.3 & 9.8 \\
FCAF3D & 47.6 & 19.4 & 30.7 & 8.8 & 89.1 & 56.9 & 73.1 & 35.2 & \textbf{90.2} & 42.4 & \textbf{63.7} & \textbf{18.5} \\ \hline \hline
Ours (anchor-based) & 57.1 & 14.9 & 11.2 & 1.3 & \textbf{97.8} & \textbf{76.5} & 65.9 & 43.2 & 88.7 & 42.4 & 40.7 & 14.4 \\
Ours (anchor-free) & \textbf{66.7} & \textbf{27.3} & \textbf{30.9} & \textbf{11.5} & 96.3 & 69.9 & 85.2 & 59.9 & 89.2 & \textbf{42.9} & 55.5 & 18.4 \\ \hline
\end{tabular}
}
\vspace{-0.1in}
\caption{\textbf{Quantitative Comparison.} Our results are reported on the VGG19 backbone. AR refers to the recall score at the specified IoU threshold, instead of the average recall.}
\vspace{-0.15in}
\label{tab:external-comparison}
\end{table*}

%% file: tabs/tab-reg_loss.tex
\begin{table}[]
\centering
\resizebox{0.90\linewidth}{!}{
\begin{tabular}{llllll}
\hline
\multirow{2}{*}{Methods} & \multirow{2}{*}{Loss} & \multicolumn{2}{c}{Recall} & \multicolumn{2}{c}{AP} \\
 &  & 0.25 & 0.50 & 0.25 & 0.50 \\ \hline
\multirow{3}{*}{Anchor-based} & Smooth $L_1$ & \textbf{98.5} & 63.2 & 51.8 & 26.6 \\
 & IoU & \textbf{98.5} & \textbf{71.3} & \textbf{61.6} & \textbf{36.7} \\
 & DIoU & 97.1 & \textbf{71.3} & 59.5 & 32.8 \\ \hline \hline
\multirow{3}{*}{Anchor-free} & Smooth $L_1$ & 96.3 & 56.6 & 76.5 & 39.9 \\
 & IoU & 96.3 & 62.5 & \textbf{78.7} & \textbf{41.0} \\
 & DIoU & \textbf{97.1} & \textbf{64.0} & 77.4 & 40.2 \\ \hline
\end{tabular}
}\vspace{-0.05in}
\caption{\textbf{Ablation results of the bounding box regression loss.}}\vspace{-0.1in}
\label{tab:reg_loss}
\end{table}

%% file: figs/fig-scene_editing.tex
\begin{figure}[t]
    \centering
    \includegraphics[width=0.49\linewidth]{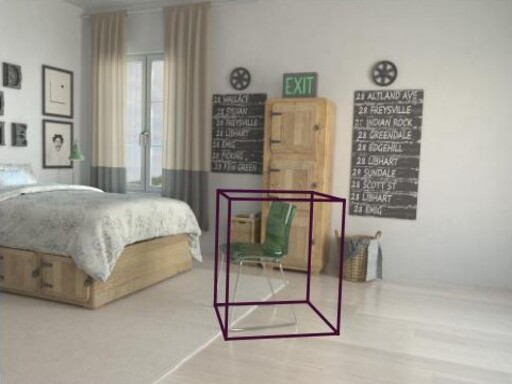}\hfill
    \includegraphics[width=0.49\linewidth]{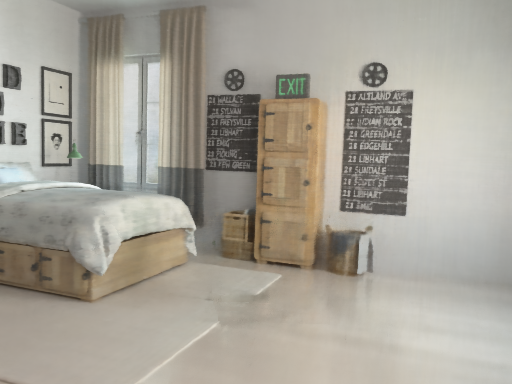}
    \vspace{-0.1in}
    \caption{\textbf{Application: Scene Editing.} 
    Removing an object in a bounding box proposed by our NeRF-RPN. }
    \vspace{-0.1in}
    \label{fig:edit}
\end{figure}

%% file: figs/fig-external-comparison.tex
\renewcommand\width{0.00cm}
\renewcommand\height{0.01cm}
\newcommand{\gt}{Ground truth}
\newcommand{\ours}{Ours}
\newcommand{\imvoxelnet}{ImVoxelNet}
\newcommand{\fcaf}{FCAF3D}

\begin{figure}[ht]
    \centering
    \captionsetup[subfloat]{labelformat=empty}

    \subfloat{\includegraphics[width = 0.24\linewidth]{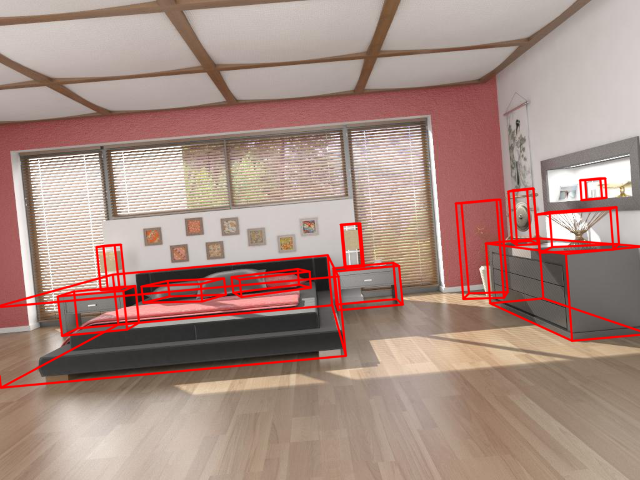}}\hspace{\width}
    \subfloat{\includegraphics[width = 0.24\linewidth]{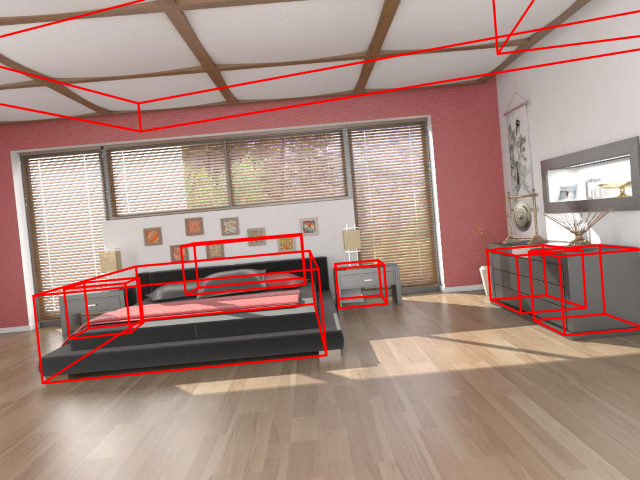}}\hspace{\width}
    \subfloat{\includegraphics[width = 0.24\linewidth]{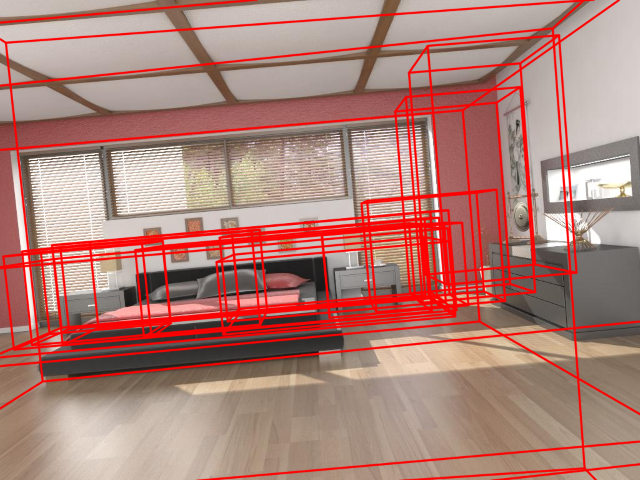}}\hspace{\width}
    \subfloat{\includegraphics[width = 0.24\linewidth]{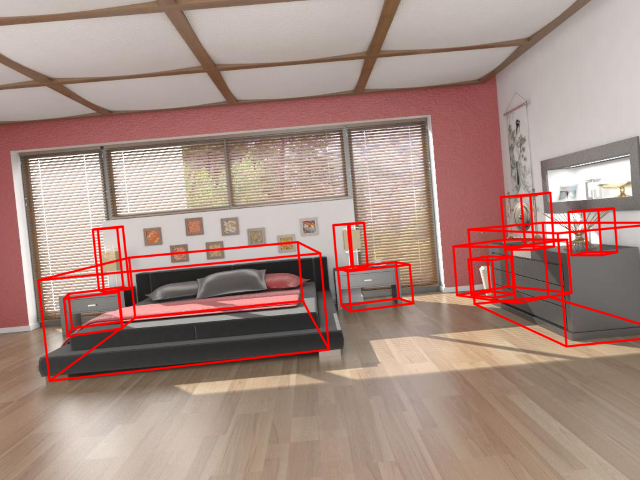}}
	
    \vspace{\height}

    \subfloat{\includegraphics[width = 0.24\linewidth]{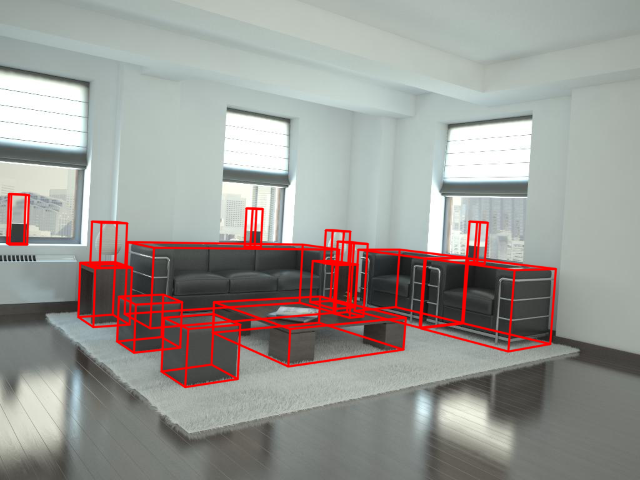}}\hspace{\width}
    \subfloat{\includegraphics[width = 0.24\linewidth]{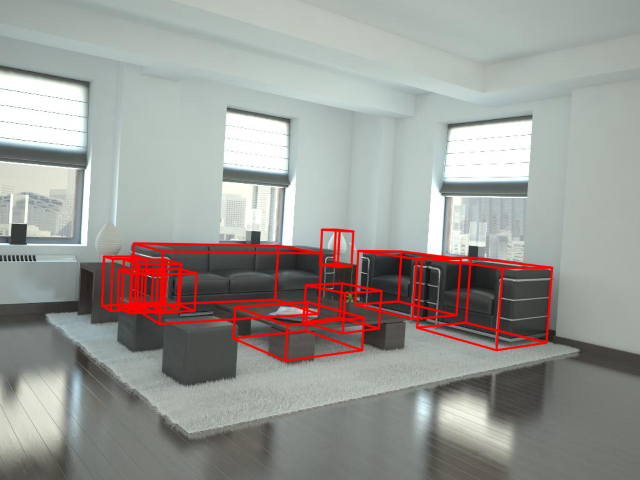}}\hspace{\width}
    \subfloat{\includegraphics[width = 0.24\linewidth]{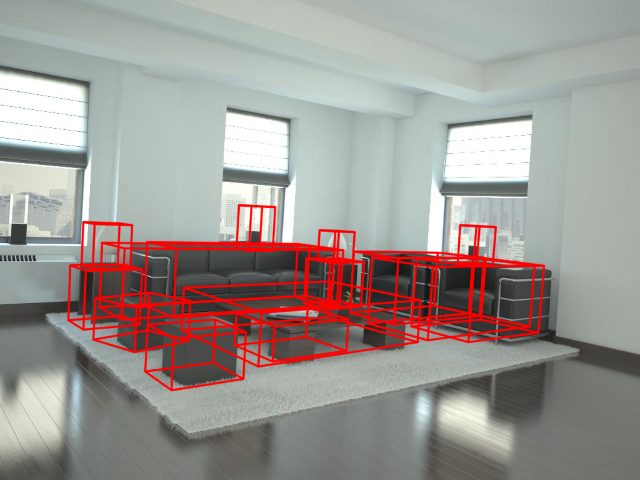}}\hspace{\width}
    \subfloat{\includegraphics[width = 0.24\linewidth]{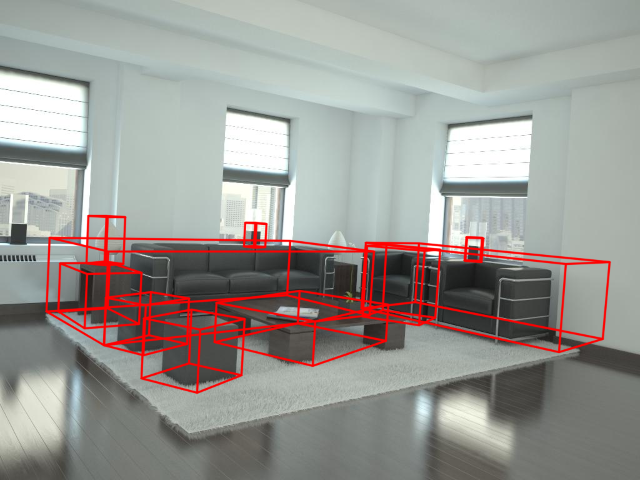}}
	
    \vspace{\height}

    \subfloat{\includegraphics[width = 0.24\linewidth]{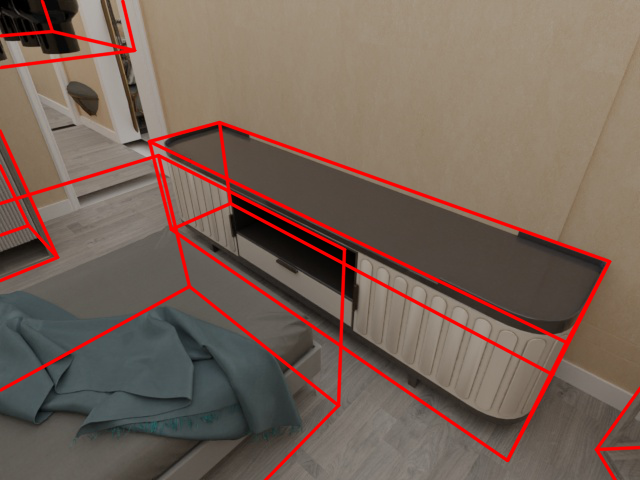}}\hspace{\width}
	\subfloat{\includegraphics[width = 0.24\linewidth]{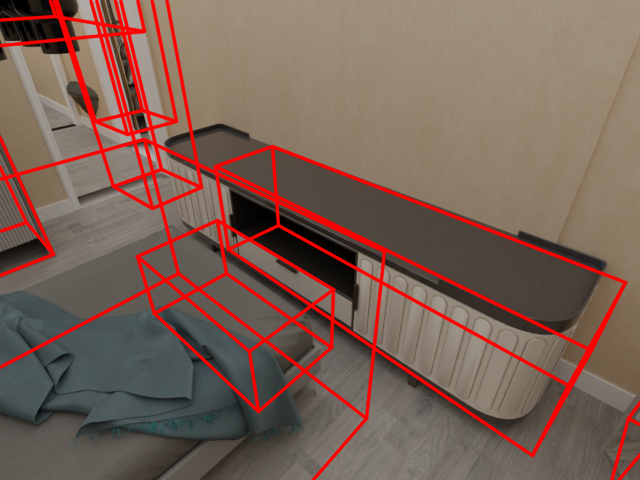}}\hspace{\width}
    \subfloat{\includegraphics[width = 0.24\linewidth]{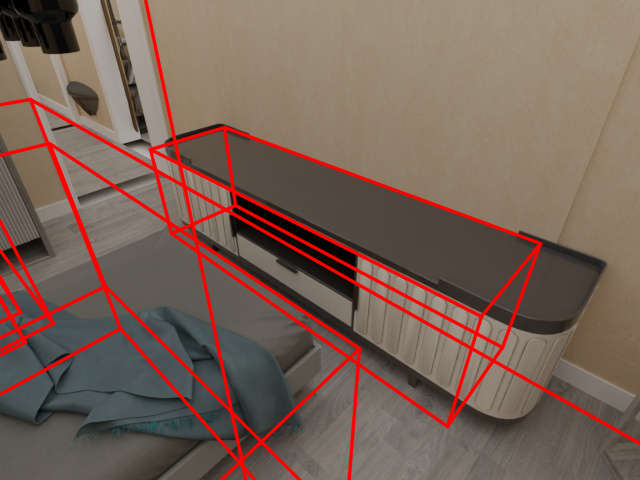}}\hspace{\width}
    \subfloat{\includegraphics[width = 0.24\linewidth]{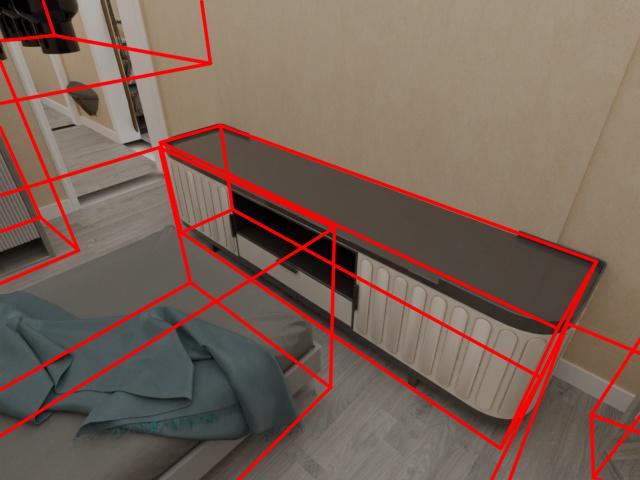}}
	
    \vspace{\height}
 
    \subfloat[\gt]{\includegraphics[width = 0.24\linewidth]{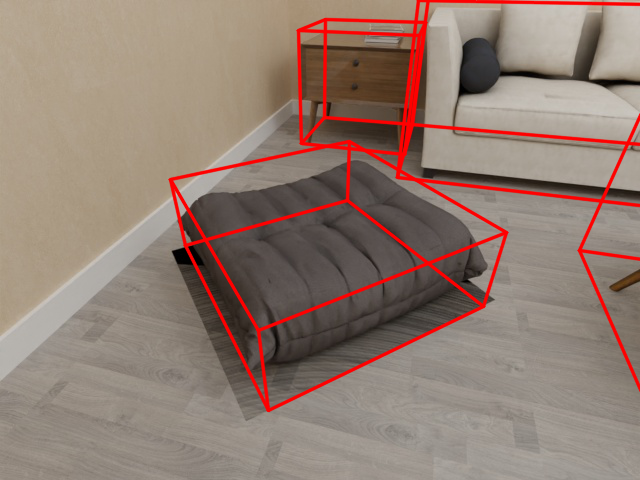}}\hspace{\width}
    \subfloat[\imvoxelnet]{\includegraphics[width = 0.24\linewidth]{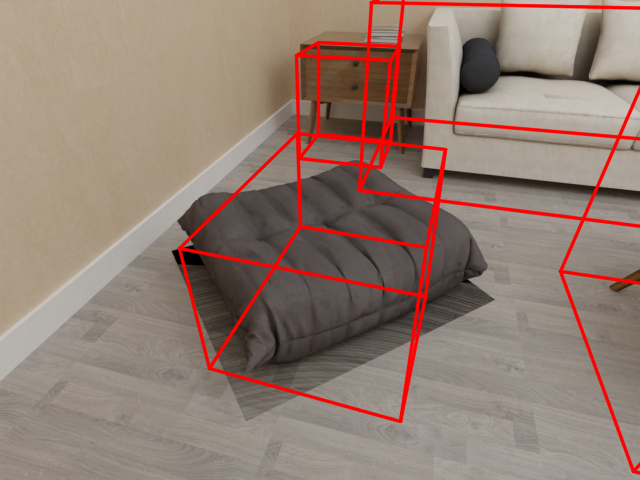}}\hspace{\width}
    \subfloat[\fcaf]{\includegraphics[width = 0.24\linewidth]{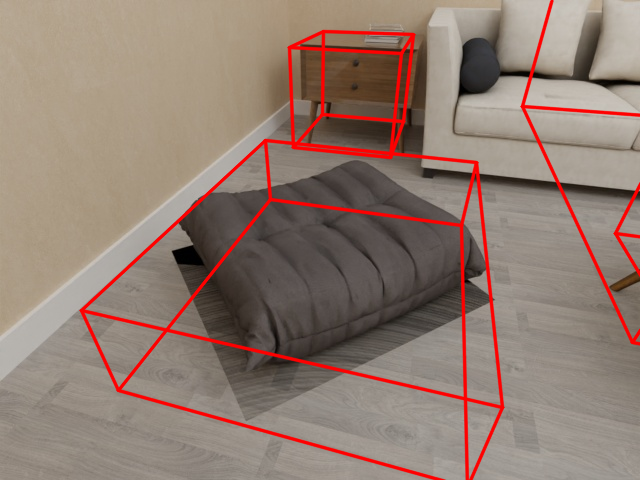}}\hspace{\width}
    \subfloat[\ours]{\includegraphics[width = 0.24\linewidth]{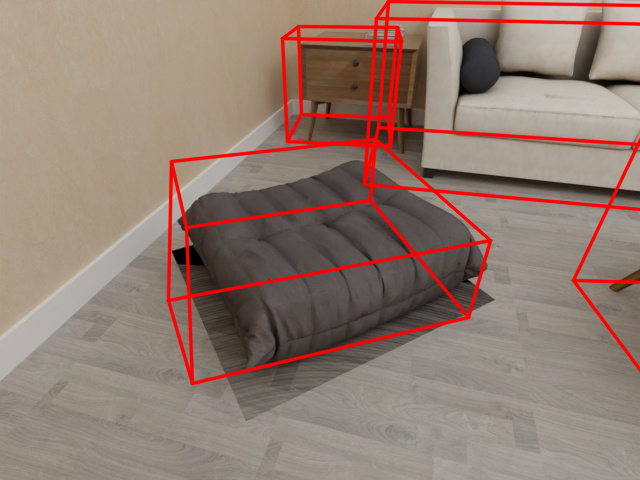}}
	
    \vspace{-0.1in}
    \caption{\textbf{Qualitative Comparison.} The first two rows are from Hypersim and the rest are from 3D-FRONT.}
    \vspace{-0.1in}
    \label{fig:external-comparison}
\end{figure}

%% file: 7_conclusions.tex
\section{Conclusion}
We propose the first significant 3D object detection framework for \nerf, NeRF-RPN, which operates on the voxel representation extracted from \nerf. By performing comprehensive experiments on different backbone networks, namely, VGG, ResNet, Swin Transformer along with anchor-based, anchor-free RPN heads, and multiple loss functions, we validate our NeRF-RPN can regress high-quality boxes directly from \nerf, without rendering images from \nerf in any view. To facilitate future work on 3D object detection in \nerf, we built a new benchmark dataset consisting of both synthetic and real-world data, with high \nerf reconstruction quality and careful bounding box labeling and cleaning. We hope NeRF-RPN will become a good baseline that can inspire and enable future work on 3D object detection in \nerfs.

%% file: tabs/tab-ablation_sampling.tex
\begin{table}[]
\centering
\resizebox{0.70\linewidth}{!}{
\begin{tabular}{lllll}
\hline
\multirow{2}{*}{Methods} & \multicolumn{2}{c}{Recall} & \multicolumn{2}{c}{AP} \\
 & \multicolumn{1}{c}{0.25} & \multicolumn{1}{c}{0.50} & \multicolumn{1}{c}{0.25} & \multicolumn{1}{c}{0.50} \\ \hline
Density only & 95.6 & \textbf{82.4} & \textbf{87.9} & \textbf{71.7} \\
Fixed directions & \textbf{96.3} & 77.9 & 84.1 & 66.3 \\
All cameras & 95.6 & 75.7 & 86.4 & 64.0 \\
Filtered cameras & \textbf{96.3} & 71.3 & 86.5 & 62.1 \\
SH coefficients & 95.6 & 69.9 & 83.2 & 57.3 \\ \hline
\end{tabular}
}
\caption{Ablation results of NeRF sampling methods. Reported metrics are calculated on the top 2500 proposals after NMS. Filtered cameras refer to removing training camera views where the sample is outside of the viewing frustums.}
\label{tab:ablation_sampling}
\end{table}

%% file: tabs/tab-objectness.tex
\begin{table}[]
\centering
\resizebox{0.75\linewidth}{!}{
\begin{tabular}{lllll}
\hline
\multirow{2}{*}{Methods} & \multicolumn{2}{c}{Hypersim} & \multicolumn{2}{c}{3D-FRONT} \\
 & $\text{AP}_{25}$ & $\text{AP}_{50}$ & $\text{AP}_{25}$ & $\text{AP}_{50}$ \\ \hline
Anchor-based &   \textbf{24.6} &
\textbf{6.2} &
\textbf{51.8} &
\textbf{26.6}
\\
\textit{~+Objectness cls.} &  12.1 &
1.2 &
36.0 &
7.4\\ \hline \hline
Anchor-free &    \textbf{27.7} &
\textbf{7.7} &
\textbf{78.7} &
\textbf{41.0} 

\\ 
\textit{~+Objectness cls.} & 14.7 &
2.5 &
44.7 &
16.8\\ \hline
\end{tabular}
}
\caption{Ablation of the objectness classification component on Hypersim and 3D-FRONT, using Swin-S as the backbone.}
\label{tab:objectness}
\end{table}

%% file: tabs/tab-2dproj_loss.tex
\begin{table}[]
\centering
\resizebox{0.70\linewidth}{!}{
\begin{tabular}{lllll}
\hline
\multirow{2}{*}{Methods} & \multicolumn{2}{c}{Recall} & \multicolumn{2}{c}{AP} \\
 & 0.25 & 0.50 & 0.25 & 0.50 \\ \hline
Anchor-based & \textbf{98.5} & 63.2 & 51.8 & \textbf{26.6} \\
\textit{~+2D proj. loss} & 97.1 & \textbf{65.4} & \textbf{58.4} & 22.2 \\ \hline \hline
Anchor-free & \textbf{96.3} & \textbf{62.5} & \textbf{78.7} & 41.0 \\
\textit{~+2D proj. loss} & \textbf{96.3} & 57.4 & 78.2 & \textbf{41.3} \\ \hline
\end{tabular}
}\vspace{-0.1in}
\caption{Ablation of the 2D projection loss run on 3D-FRONT, using Swin-S as the backbone.}\vspace{-0.2in}
\label{tab:proj_loss}
\end{table}